\documentclass[journal]{IEEEtran}

\usepackage{amssymb}
\usepackage{amsthm}
\usepackage{amsmath}
\usepackage{lineno}
\usepackage{graphicx}
\usepackage{booktabs}
\usepackage{hyperref}
\usepackage{threeparttable}
\usepackage{colortbl}
\usepackage{graphbox}
\usepackage[export]{adjustbox}
\usepackage{subcaption}
\usepackage{comment}
\usepackage{multirow}
\usepackage{textcomp}
\usepackage{orcidlink}

\newcommand{\GP}{\mathcal{GP}}

\ifCLASSINFOpdf
\else
\fi
\begin{document}

\title{Daily Land Surface Temperature Reconstruction in Landsat Cross-Track Areas Using Deep Ensemble Learning With Uncertainty Quantification}

\author{Shengjie~Liu\,\orcidlink{0000-0003-0253-7410},~\IEEEmembership{Student Member,~IEEE,} 
        Siqin~Wang, 
        and~Lu~Zhang
\thanks{Preprint. S.L. was supported in part by the USC Dana and David Dornsife College of Letters, Arts and Sciences/Graduate School Fellowship, and in part by the National Institute of Environmental Health Sciences of the National Institutes of Health (NIH) through the Southern California Environmental Health Sciences Center under Grant P30ES007048. L.Z. was supported in part by NIH under Grant P30ES007048 and P20HL176204. The views and conclusions expressed in this study are those of the authors and do not necessarily represent the official policies or endorsements of the NIH.} 
\thanks{S. Liu and S. Wang are with Spatial Sciences Institute, Dornsife College of Letters, Arts and Sciences, University of Southern California, Los Angeles, CA 90089, USA (e-mail: skrisliu@gmail.com; siqinwan@usc.edu).}
\thanks{L. Zhang is with Division of Biostatistics and Health Data Science, Department of Population and Public Health Sciences, Keck School of Medicine, University of Southern California, Los Angeles, CA 90089, USA (e-mail: lzhang63@usc.edu).}}

\markboth{Preprint.}%
{Preprint.}

\maketitle

\begin{abstract}
Many real-world applications rely on land surface temperature (LST) data at high spatiotemporal resolution. In complex urban areas, LST exhibits significant variations, fluctuating dramatically within and across city blocks. Landsat provides high spatial resolution data at 100 meters but is limited by long revisit time, with cloud cover further disrupting data collection. Here, we propose DELAG, a deep ensemble learning method that integrates annual temperature cycles and Gaussian processes, to reconstruct Landsat LST in complex urban areas. Leveraging the cross-track characteristics and dual-satellite operation of Landsat since 2021, we further enhance data availability to 4 scenes every 16 days. We select New York City, London and Hong Kong from three different continents as study areas. Experiments show that DELAG successfully reconstructed LST in the three cities under clear-sky (RMSE = 0.73-0.96 K) and heavily-cloudy (RMSE = 0.84-1.62 K) situations, superior to existing methods. Additionally, DELAG can quantify uncertainty that enhances LST reconstruction reliability. We further tested the reconstructed LST to estimate near-surface air temperature, achieving results (RMSE = 1.48-2.11 K) comparable to those derived from clear-sky LST (RMSE = 1.63-2.02 K). The results demonstrate the successful reconstruction through DELAG and highlight the broader applications of LST reconstruction for estimating accurate air temperature. Our study thus provides a novel and practical method for Landsat LST reconstruction, particularly suited for complex urban areas within Landsat cross-track areas, taking one step toward addressing complex climate events at high spatiotemporal resolution. Code and data are available at \url{https://skrisliu.com/delag}.
\end{abstract}


This is the preprint version. For the final version, go to \href{https://doi.org/10.1109/tgrs.2025.3643985}{https://doi.org/10.1109/tgrs.2025.3643985}.

\IEEEpeerreviewmaketitle

\section{Introduction}
Continuous land surface temperature (LST) data from satellites is fundamental to achieve sustainable development, to understand climate, and to analyze climate impacts on planetary health~\cite{burke2021using}. Many of the real-world applications rely on satellite LST data, including monitoring wildfires, weather forecasting, and understanding temperature-related mortality~\cite{shao2022constructing, shi2015impacts, powers2017weather}. Although satellites can provide LST data on a large scale, these data are often incomplete due to cloudy weather conditions and sometimes sensor failure~\cite{weiss2014effective, militino2019interpolation, malakar2018operational}. For high spatial resolution satellite data, due to sensor technology, the trade-off between spatial and temporal resolutions further restricts daily data collection. Although daily LST data is crucial for real-world applications, currently, we don't have good enough daily LST data at high spatiotemporal resolution~\cite{li2020adjustment}.

\begin{figure*}[!t]
\centering
\includegraphics[width=.92\linewidth]{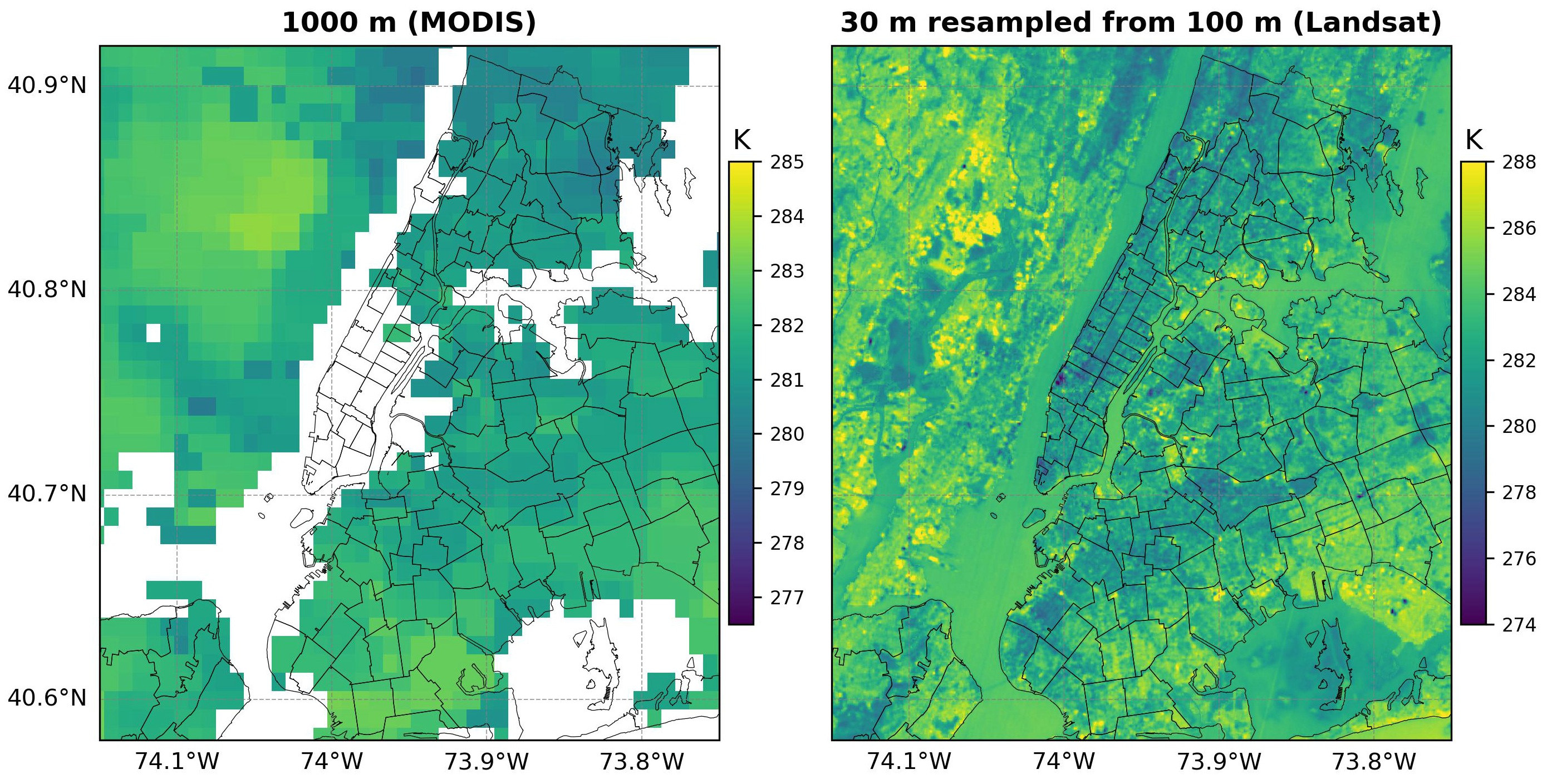}
\caption{Spatial resolution comparison of satellite-derived land surface temperature products in New York City: MODIS (1~km), Landsat (30~m resampled from 100~m). The overlaid polygons are the New York City merged zip-code boundaries that have daily counts of emergency department visit data. The coarse-resolution LST data is not sufficient to capture spatial variations and urban heterogeneity. Landsat LST can capture temperature variations but is limited by temporal frequency and clouds.}
\label{fig:ComparisonLSTintro}
\end{figure*}

The most commonly used temperature data today is the MODIS observational, gridMET hybrid and ERA5 reanalysis data, with 1~km, 4~km and 11~km spatial resolution, respectively~\cite{abatzoglou2013development, wang2022evaluating}. But, at their kilometer resolution, it is impossible to capture temperature heterogeneity within a city~\cite{wu2021spatially}. In complex urban areas, temperature changes dramatically over short distances, ranging from a few meters to street blocks~\cite{chen2020evaluating}. For example, in New York City as shown in Fig.~\ref{fig:ComparisonLSTintro}, many urban parks have been able to reduce daytime temperature with vegetation. The city has complex street blocks exhibiting large temperature variations, but these variations are not captured from the 1~km resolution MODIS data (Fig.~\ref{fig:ComparisonLSTintro}). The coarse-resolution LST data and the resulting coarse-resolution water masks also make data unavailable in many coastal urban areas with high population density such as Manhattan (Fig.~\ref{fig:ComparisonLSTintro}). Having fine-scale LST data is therefore critical due to the complex landscape, vast temperature heterogeneity, and high population concentration in cities that are more vulnerable to climate impacts~\cite{broto2013survey}. 
Additionally, in real-world applications, many datasets are aggregated to small geographical units, such as zip codes (Fig.~\ref{fig:ComparisonLSTintro}). Many socioeconomic and health datasets are available at this level to support research on various topics, including air pollution, respiratory diseases, and temperature-related mortality~\cite{gasparrini2022small, luo2022long, liu2024effects}. However, the development of high resolution temperature data lacks behind, and many real-world applications had to settle to the coarse-resolution LST data~\cite{yazdi2021long, dahl2019increased, jiao2023analysis}. 

In order to have daily LST data for real-world applications, many efforts have been devoted for LST data reconstruction~\cite{wu2021spatially, zeng2018two, xiao2023integrated, you2024reconstruction}. These methods are often applied to the MODIS, VIIRS and Himawari-8 LST data that have a spatial resolution of 1-10~km~\cite{wu2021spatially,chen2023stepwise}. The annual temperature cycle (ATC) model, for example, is one of the popular methods~\cite{xia2021modeling,quan2023generating}. The ATC model assumes that daily mean temperature follows a cosine function with three parameters, and the three parameters have their physical meaning, representing the annual mean temperature, the amplitude, and the phase shift that are critical to understand changes over time~\cite{stine2009changes}. However, the ATC model cannot capture daily fluctuation. One direction is to revise the three-parameter ATC model to an enhanced four-parameter ATC model to incorporate reanalysis temperature data~\cite{hong2021simple} or phenology information~\cite{xia2021modeling, yang2024annual}. Another solution is to use Gaussian processes to model the residuals between the observed and the ATC-reconstructed LST data~\cite{fu2016consistent}. These existing methods are often applied to LST data reconstruction at the kilometer resolution due to data availability.

Landsat is the only widely-used satellite series with sufficiently high spatial resolution to capture temperature heterogeneity within a city. Landsat has a revisit time of 16 days, an often-used excuse (in addition to clouds) to eliminate its potential to generate daily LST data. In the existing literature, only a very small subset has explored Landsat LST reconstruction. For example, Zhu~\emph{et~al.}~\cite{zhu2022reconstruction} proposed an ATC-based novel method to reconstruct cloudy LST based on a series of procedures including cloud removal, shadow identification, and grouping from spatially adjacent similar pixels. An earlier work used temporal Gaussian processes for daily Landsat LST reconstruction without considering spatial correlation~\cite{fu2016consistent}. As a restriction of Landsat's long revisit time, both studies adopted an inter-annual approach to ensure sufficient data support, making the interpretation of ATC parameters that are supposed to have physical meaning difficult. These methods are also limited to dates with Landsat observations; on many dates without Landsat observations, the uncertainty drawn from these approaches can be large, but the reliability and uncertainty of the reconstruction are unknown. 

\begin{figure*}[!t]
\centering
\includegraphics[width=.99\linewidth]{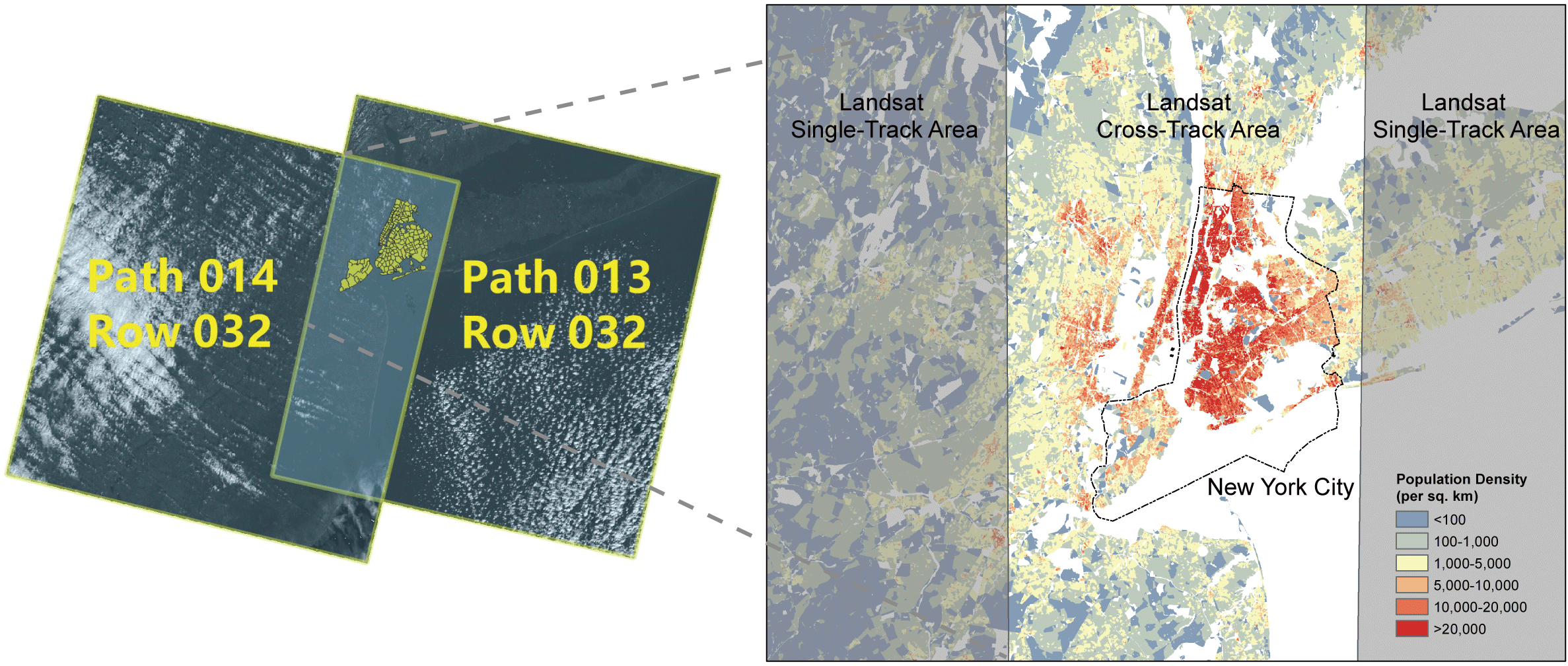}
\caption{New York City is situated within the Landsat cross-track areas, completed being covered by two Landsat scenes (Path 014 Row 032, Path 013 and Row 032). This area, New York City and the surroundings, has complex urban landscape and high population density.}  
\label{fig:StudyAreaNYC}
\end{figure*}

To generate high resolution LST data on a daily basis, especially in cities, we here propose a framework that consists of two parts: from the perspectives of both data and method. From the data perspective, we leverage the Landsat cross-track areas and highlight their overlooked role in LST reconstruction. Although Landsat has a revisit time of 16 days, the Landsat cross-track areas (2 observations per 16 days) cover many cities that can benefit from this special setting. For example, the densely-populated New York City is completely situated within two Landsat tracks~(Fig.~\ref{fig:StudyAreaNYC}). Two other cities---London and Hong Kong---in this study also have their majority within the cross-track areas. With both Landsat 8 and 9 satellites in operation since 2021, the frequency is further enhanced to 4 scenes per 16 days, making daily Landsat LST reconstruction for practical applications toward feasible. From the method perspective, in this study, we propose DELAG, a deep ensemble learning method incorporating enhanced ATC models and Gaussian processes for Landsat LST reconstruction. DELAG utilizes deep ensembles to achieve prediction intervals---the first of its kind. The enhanced ATC model and Gaussian processes are used to link observed pixels to cloud-covered pixels. There is no LST site within the study areas, and given that many real-world applications require air temperature (instead of LST) derived from satellites~\cite{kloog2015using} and the ability of LST to generate accurate air temperature~\cite{hooker2018global}, we validate the proposed method by comparing the air temperature generated from observed LST and reconstructed LST using a simple linear statistical method, a validation approach commonly used in the literature~\cite{yang2024annual, zhao2020reconstruction}. 

In the remainder of this paper, we first provide a concise literature review on the current LST reconstruction methods in Section~\ref{sec:literature}. We then introduce and visualize the Landsat cross-track areas in Section~\ref{sec:LandsatCrossTrack}. We show that Landsat cross-track areas are sufficiently generalized, in many countries and regions, especially in mid- to high-latitude densely populated areas. We select three representative cities---New York City, London and Hong Kong---across three different continents as study areas here to demonstrate the generalization of the proposed framework. We then detail the proposed method (DELAG), the auxiliary data and validation strategies in Section~\ref{sec:DataMethod}. In Section~\ref{sec:result}, we show the experimental results in the three cities, the ablation analysis, and the indirect validation (and application) with near-surface air temperature data. We discuss the broader impacts and conclude this study in Section~\ref{sec:ConclusionDiscussion}. 

\section{Literature on LST Reconstruction Methods}
\label{sec:literature}
Reconstruction of LST data, from an application perspective, can be broadly categorized into two types: all-weather (at least daily) reconstruction and non-all-weather reconstruction. Based on this distinction, all-weather approaches can be further classified according to the sources or channels through which they obtain information about unobserved LST. These methods include spatial and/or temporal interpolation and their variants, data fusion integrating multiple satellites, passive microwave and ground-based observations, surface energy balance modeling, and numerical modeling of Earth system processes. Two comprehensive reviews of LST reconstruction methods are available in the literature~\cite{wu2021spatially, jia2024advances}. In this section, we primarily provide a focused overview relevant to daily LST reconstruction, with particular attention to ATC models, surface energy balance, and deep learning approaches. The proposed method can be understood as a hybrid approach that integrates information derived from spatiotemporal interpolation via ATC and Gaussian processes and surface energy balance through the output of numerical modeling (ERA5).

\subsection{Spatial and Temporal Interpolation With Land Surface Information}
Spatial interpolation aims to gap-fill LST data using valid observations available at the same time point. In its narrow definition, this class of methods explicitly models spatial proximity---typically Euclidean distance---between observed and unobserved pixels, often employing techniques such as Gaussian processes and graph-based models~\cite{rolland2024improving}. More broadly, the notion of distance is extended to kernel-based or feature-based distances (also referred to as latent distances), where similarity is measured through spectral reflectance including vegetation index or latent representations of land surface characteristics~\cite{zhu2022reconstruction, xia2019combining, liu2024deep}. Some studies further shift the problem from interpolation to a regression framework, where the objective becomes learning a mapping from input features \( x \) to LST \( T_s = f(x) \), often using machine learning models~\cite{you2024reconstruction, yang2024annual}. By redefining distance in terms of feature similarity rather than geographic location, these models relax the constraint of spatial adjacency and are more adaptable to real-world conditions, where cloud cover is typically non-random and frequently occludes large contiguous regions.

Regardless of whether spatial interpolation is defined narrowly or broadly, its effectiveness depends on the availability of valid observations. On heavily overcast or rainy days, zero valid observations are not uncommon, and this challenge is compounded by the infrequent revisit times of many satellite sensors. Consequently, spatial interpolation methods are often applied to MODIS data, which provides partial daily observations~\cite{you2024reconstruction, yang2019integrated, kang2018reconstruction}. To address data gaps more effectively, temporal interpolation---typically based on the ATC model---is frequently integrated prior to spatial methods~\cite{zhu2022reconstruction, zhang2021global, bian2024integrated, ding2022reconstruction}. Still, these combined approaches are limited on days without any available observations.

\subsection{Data Fusion and Super-Resolution}
Integrating observations from multiple data sources is a natural step toward improving data availability and reliability~\cite{zhang2023multiinformation, lee2025guided, duan2025enhanced}. With the growing interest in data fusion and super-resolution techniques in machine learning, these methods have become increasingly popular for LST reconstruction~\cite{lee2025guided, weng2014generating, tang2024generation}. In a typical machine learning framework, given a pair of low- and high-resolution data, the goal is to find a function \( f(\cdot) \) to map low-resolution data to high resolution, i.e., \( x_{\text{high}} = f(x_{\text{low}}) \). For example, several spatiotemporal fusion methods have been applied to downscale GOES-16 LST data from 2~km to Landsat's 100~m resolution~\cite{tang2024generation}. Data fusion requires cloud-free pairs of low- and high-resolution images, which can be a limiting factor in real-world applications. To address this, Li~\emph{et al.}~\cite{li2025lfsr} first filled gaps in the low-resolution image using a clear reference image before applying a super-resolution neural network. While data fusion and super-resolution methods hold strong potential due to their data-driven nature, they remain limited by weather conditions and typically require more intensive data collection and training resources compared to other approaches. In some cases, low-resolution images may not be available at all. Microwave observations, which are not affected by cloud cover, offer some promise for obtaining all-weather LST data, but their current temporal frequency is still insufficient for daily reconstruction~\cite{li2024downscaling}.

\subsection{Surface Energy Balance-Based Approaches}
As the availability of observational data fundamentally limits methods based on interpolation and data fusion, a growing direction is to supplement these approaches with information derived from physical principles, particularly surface energy balance~\cite{tan2021reconstruction}. This additional step can take relatively simple forms, such as applying a linear correction to adjust for cloud effects~\cite{tan2021reconstruction}, or comparing cloud-contaminated pixels with clear-sky reference images using similar-pixel techniques~\cite{yu2019effective}. However, due to the complexity of heat transfer processes and surface energy balance calculations at fine spatial scales, such modeling is often oversimplified, and the representation of cloud effects within these calculations remains insufficient~\cite{jia2024advances}. Numerical models and climate reanalysis data, which explicitly incorporate surface energy balance into calculations at the global scale, produce outputs that implicitly encode this physical information. Building on this foundation, recent studies have begun integrating numerical model outputs to estimate the difference between observed and cloud-covered LST as simulated by ERA5~\cite{xu2022reconstructing}. These approaches involve first estimating the clear-sky and cloud-cover LST difference using ERA5 data, then applying the result to higher-resolution observational data for reconstruction~\cite{xu2022reconstructing, liu2023estimating, li2025downscaling}.

\subsection{Linking Prior Work to the Proposed Method}
The proposed method is primarily a hybrid approach that combines spatiotemporal interpolation (via ATC and Gaussian Processes) with information from surface energy balance, incorporated as a linear term based on coarse-resolution ERA5 data that is the output of numerical models. It differs from prior work by integrating these components into a single equation and treating ERA5 not as a spatially resolved image but as the areal mean, an assumption that, when combined with observational data from Landsat, enables reconstruction at the Landsat scale. Additional contributions of the proposed method include uncertainty quantification through deep ensemble learning, as well as a broader framework for leveraging Landsat cross-track areas to increase data frequency. Finally, we introduce a validation strategy based on near-surface air temperature data. While not entirely new, this approach is less commonly used than traditional in-situ LST validation that is infeasible in major metropolitan areas like the three major cities in this study.

\section{Landsat Cross-Track Areas}
\label{sec:LandsatCrossTrack}
The Landsat satellites are a series of polar-orbiting satellites that are designed for land use and land cover monitoring~\cite{irons2012next}. Since Landsat 4 launched in 1982, the Landsat satellites have thermal sensors onboard that can record land surface temperature. The current operating satellites (Landsat 8 and 9) have two thermal bands at 10.6--11.2~\textmu m (Band 10) and 11.5--12.5~\textmu m (Band 11). Because of polar orbiting, each Landsat satellite has a revisit time of 16 days. Between two adjacent Landsat tracks (recording data on two different dates), there is an overlapped area that has dual observations~\cite{kovalskyy2013global}. The cross-track characteristics are often overlooked, as these areas are commonly assumed to be insignificant. However, they actually cover a considerable area, especially in mid- to high-latitude regions, including some densely populated urban areas. Additionally, from a data perspective, we have the potential to recover the entire LST surface on Earth from the partial observations. 

\begin{figure}[!t]
\centering
\includegraphics[width=0.9\linewidth]{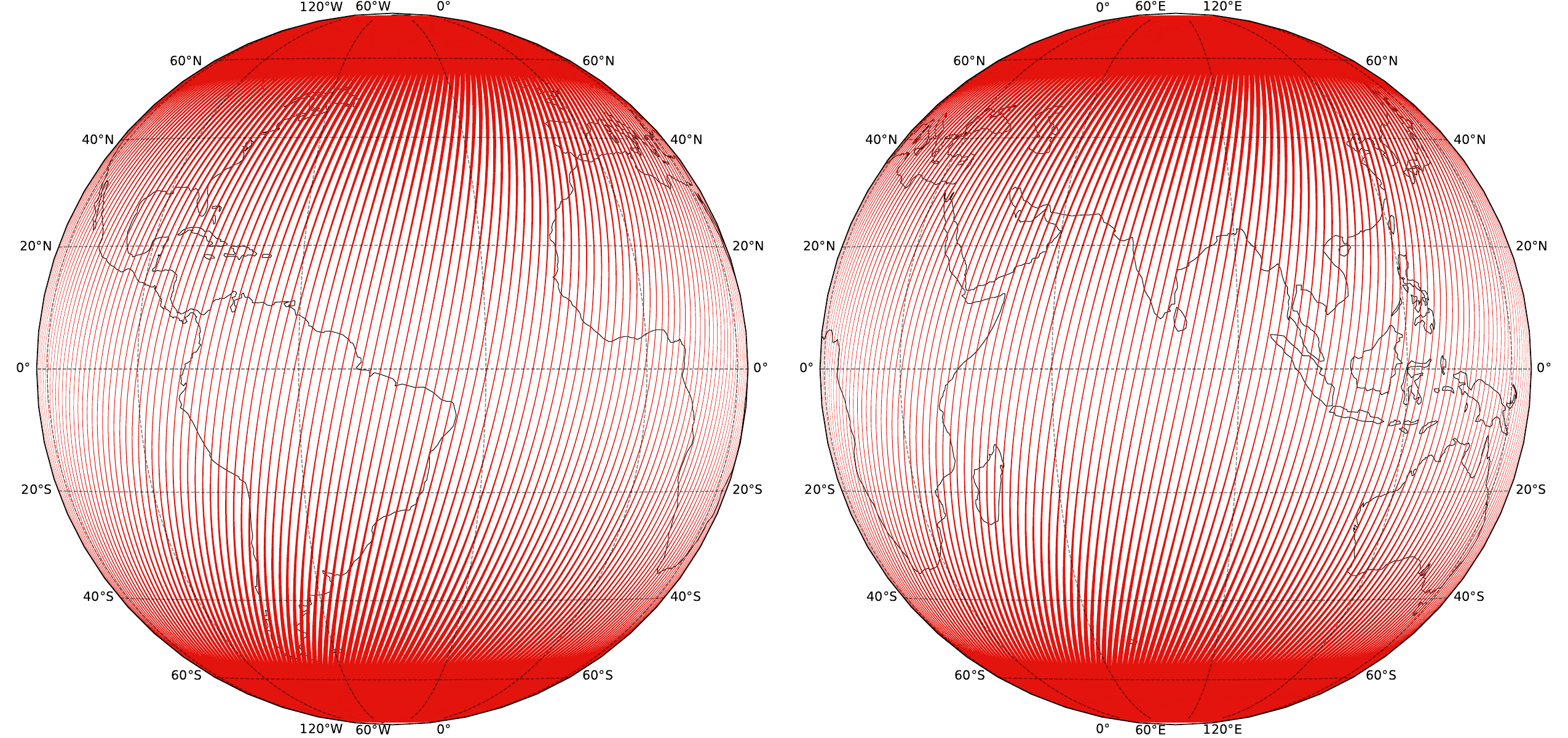}
\caption{Global areas covered by two or more Landsat scenes (in red) within a revisit period of 16 days}
\label{fig:overlapTrack}
\end{figure}

\begin{figure*}[!t]
\centering
\includegraphics[width=0.9\linewidth]{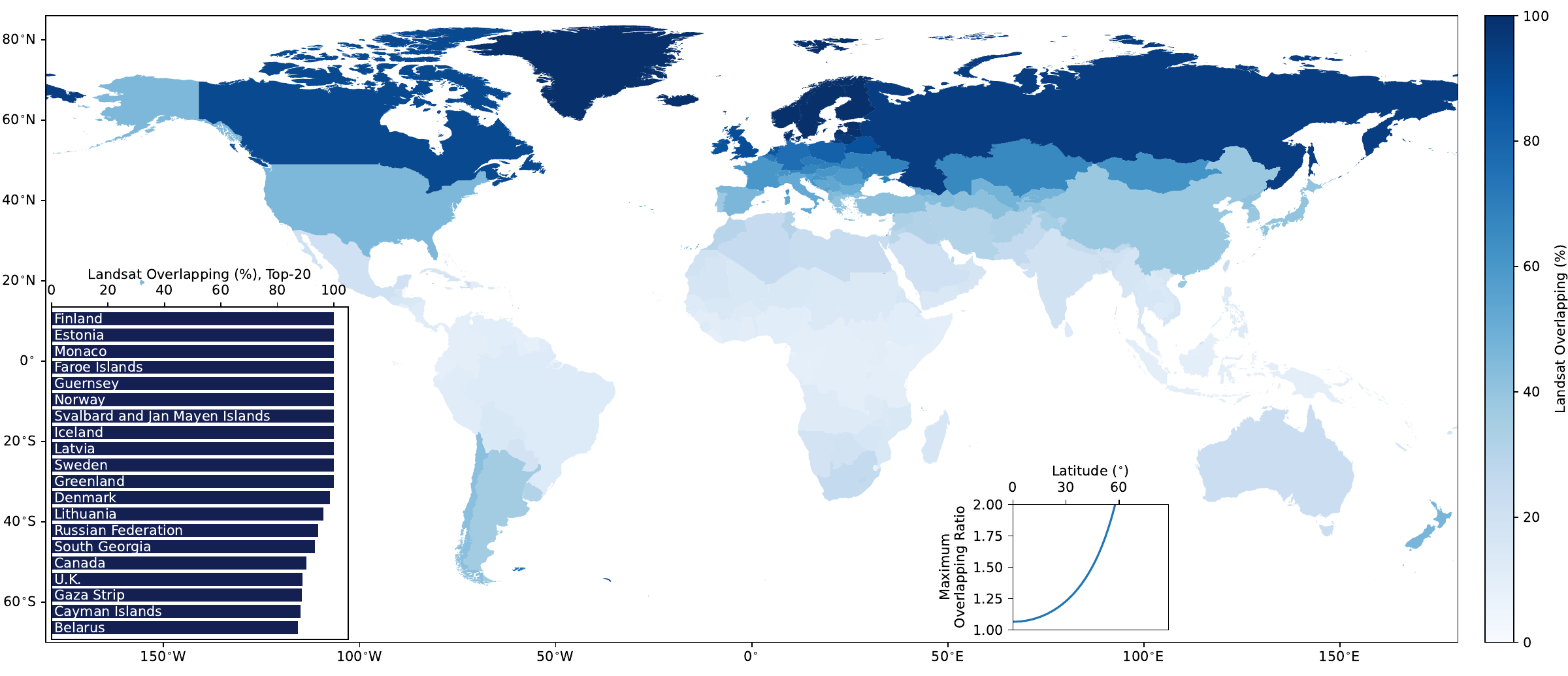}
\caption{Percentage of Landsat cross-track areas in global countries and regions. The top-20 are listed on the lower left. The ratio of Landsat coverage as a function of latitude is shown in the middle right.}
\label{fig:worldmap}
\end{figure*}

In the following calculation and visualization, we demonstrate that the Landsat cross-track areas are not as insignificant as previously assumed. In the current specification, Landsat captures each scene in a dimension of 185~km~$\times$~180~km following its polar-orbiting trajectory~\cite{irons2012next}. In Fig.~\ref{fig:StudyAreaNYC}, we show two sample scenes and highlight the cross-track areas that completely cover New York City. Landsat scenes have a width of 185~km ($L_{landsat}$) in a roughly horizontal direction, with a 8.2$^{\circ}$ shift parallel to latitude, leading to a horizontal coverage of 183.1~km. Every 16 days, a Landsat satellite travels 233 tracks, and then revisits. The ratio $\phi_{lat}$ between the horizontal length of all Landsat scenes within a revisit period (16 days, 233 tracks) and the latitude (with a length of $L_{lat}$) can be calculated as 
\begin{equation}
  \phi_{lat} =  \frac{L_{landsat} \times \cos \left(8.2^{\circ}\right) }{  L_{lat} } \times 233 \;. 
\end{equation}
The ratio reaches its minimum (1.07) at the Equator, about 7\% of the areas at the Equator are covered by two Landsat scenes per 16 days. The percentage of overlapping areas increases with latitude increasing and reaches to a ratio of 1.50 at 45$^{\circ}$ latitude: half of the areas at 45$^{\circ}$N and 45$^{\circ}$S are covered by two Landsat scenes per 16 days. We compute and visualize these Landsat cross-track areas in Fig.~\ref{fig:overlapTrack}, and the percentage of the cross-track areas in each country and region in Fig.~\ref{fig:worldmap}. 
Eleven countries and regions are 100\% located in the Landsat cross-track areas, and many countries and regions have half of their areas located in the Landsat cross-track areas. Mid-to-high-latitude countries and regions, including most in Europe, Canada, and Russia have a high percentage of cross-track areas. At the metropolitan level, the cross-track areas cover many densely-populated urban areas including New York City (100\%), London ($>$95\% areas), and Hong Kong ($>$90\% urban areas). As a combined result of cross tracks and dual satellites, we have four scenes per 16 days in these cities, making it toward feasible for daily Landsat LST reconstruction. In this study, we select New York City, London and Hong Kong as the three representative cities as study areas to demonstrate the generalization of the proposed framework and method.

\section{Study Areas and Data}
\subsection{Study Areas}
\subsubsection{New York City}
New York City (NYC), located at 40.7$^{\circ}$N and 74.0$^{\circ}$W and fully covered by two Landsat scenes (Path 014 Row 032 and Path 013 Row 032), is the most densely populated megacity in the U.S. and serves as the primary case in this study for three reasons. First, NYC's health department publicly releases daily syndromic surveillance data (e.g., respiratory emergency visits) at the zip code level. This dataset has been widely used to study the health impacts of air pollutants and other environmental exposures~\cite{liu2024effects, heffernan2004syndromic, greene2021nowcasting}. However, no daily LST data is currently available at the zip code scale, limiting our ability to assess temperature-related impacts and disparities within the city. Second, NYC lies entirely within Landsat's cross-track coverage (Fig.~\ref{fig:StudyAreaNYC}), making it an ideal study area. Within the first 16-day circle (Jan 4-19) in 2023, Landsat 8 obtained data on the 4th and 13th, and Landsat 9 obtained data on the 5th and 12th, and then the circle repeats every 16 days, totaling 92 scenes in 2023. Third, NYC's complex urban landscape and high temperature variability require fine-resolution data. With 8.3 million residents and a density of 11,000 per km$^2$, about 110 people per 100 m $\times$ 100 m Landsat thermal pixel, LST varies substantially across the city. NYC has a humid subtropical climate, with its surroundings located in the transition zone between humid subtropical and humid continental climates. The daily mean temperature is about 0.7$^{\circ}$C in January and 25.3$^{\circ}$C in July. The typical daily minimum and maximum range is between -2.3 and 29.4$^{\circ}$C.

\subsubsection{London}
To demonstrate the generalization of the proposed framework, London is selected as a second study case. Located at 51.5$^{\circ}$N on the Prime Meridian (0.0$^{\circ}$E), London is the largest city in the U.K. and Western Europe, with a population of 8.8 million across 1,572 km$^2$ (5,690 residents per km$^2$). London has a temperate oceanic climate, with daily mean temperatures ranging from 5.6 to 19$^{\circ}$C. Typical nighttime temperatures are around 2.7$^{\circ}$C, while typical daytime highs are around 23.9$^{\circ}$C and can be above 28$^{\circ}$C in summertime. Despite its higher latitude, London experiences a milder annual temperature range compared to NYC, largely due to the influence of the North Atlantic Current. With the exception of a small portion in the southeast, most of London falls within Landsat's cross-track areas (Path 202 Row 024, Path 201 Row 024). Within the first 16-day circle (Jan 9-24) in 2023, Landsat 8 obtained data on 9th and 18th, Landsat 9 obtained data on 10th and 17th, and then the circle repeats every 16 days, totaling 86 scenes in 2023. The city is divided into fine-scale statistical units (983 Middle Super Output Areas and 4,835 Lower Super Output Areas), most of which are smaller than 1 km$^2$, i.e., finer than the MODIS resolution. These geographical units have mortality and other socioeconomic and health statistics that can be used for social good, but as a result of the lack of high resolution temperature data, existing studies had to settle on using coarse resolution temperature data that limits their capability~\cite{gasparrini2022small, murage2020individual, abed2023spatial}.

\subsubsection{Hong Kong}
Hong Kong is selected as the third study area to demonstrate the generalization of the proposed framework and method across different continents and climate zones. Hong Kong has a humid subtropical climate, with a daily mean temperature between 16.5--28.9$^{\circ}$C. Typical winter average daily minimum temperature is 13$^{\circ}$C in January and average daily maximum is 31.1$^{\circ}$C in August. Similar to NYC and London, most urban areas are also situated within the Landsat cross-track areas (Path 121 Row 045, Path 122 Row 044). Within the first 16-day circle (Jan 1-16) in 2023, Landsat 9 obtained data on the 1st and 10th, Landsat 8 obtained data on the 2nd and 9th, and then the circle repeats every 16 days, totaling 91 scenes within 2023. As a coastal city in Asia, Hong Kong has a land area of 1,073 km$^2$, housing 7.5 million people (6,989 per km$^2$). The city maintains a large portion (75\%) of undeveloped lands, further intensifying population density. As a high-density and subtropical city, Hong Kong faces various extreme climate events, in need of high spatial resolution temperature data to capture local-scale variation due to urban topography, vertical morphology, and proximity to the sea. This is therefore a representative case for assessing the added value of the proposed method in a coastal, high-density, and climate-vulnerable setting.

\begin{figure}[!t]
    \centering
    \includegraphics[width=0.9\linewidth]{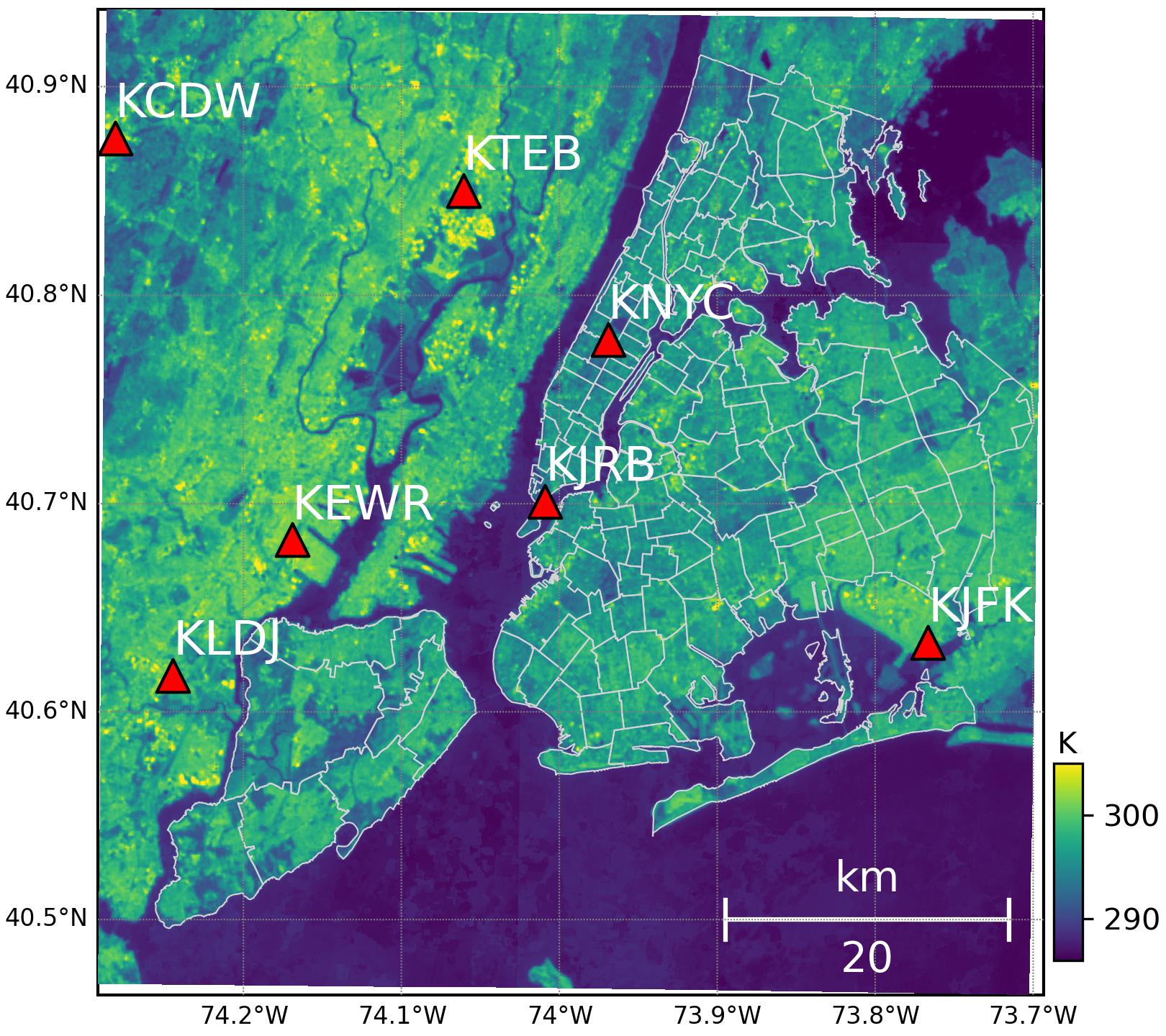}
    \includegraphics[width=0.9\linewidth]{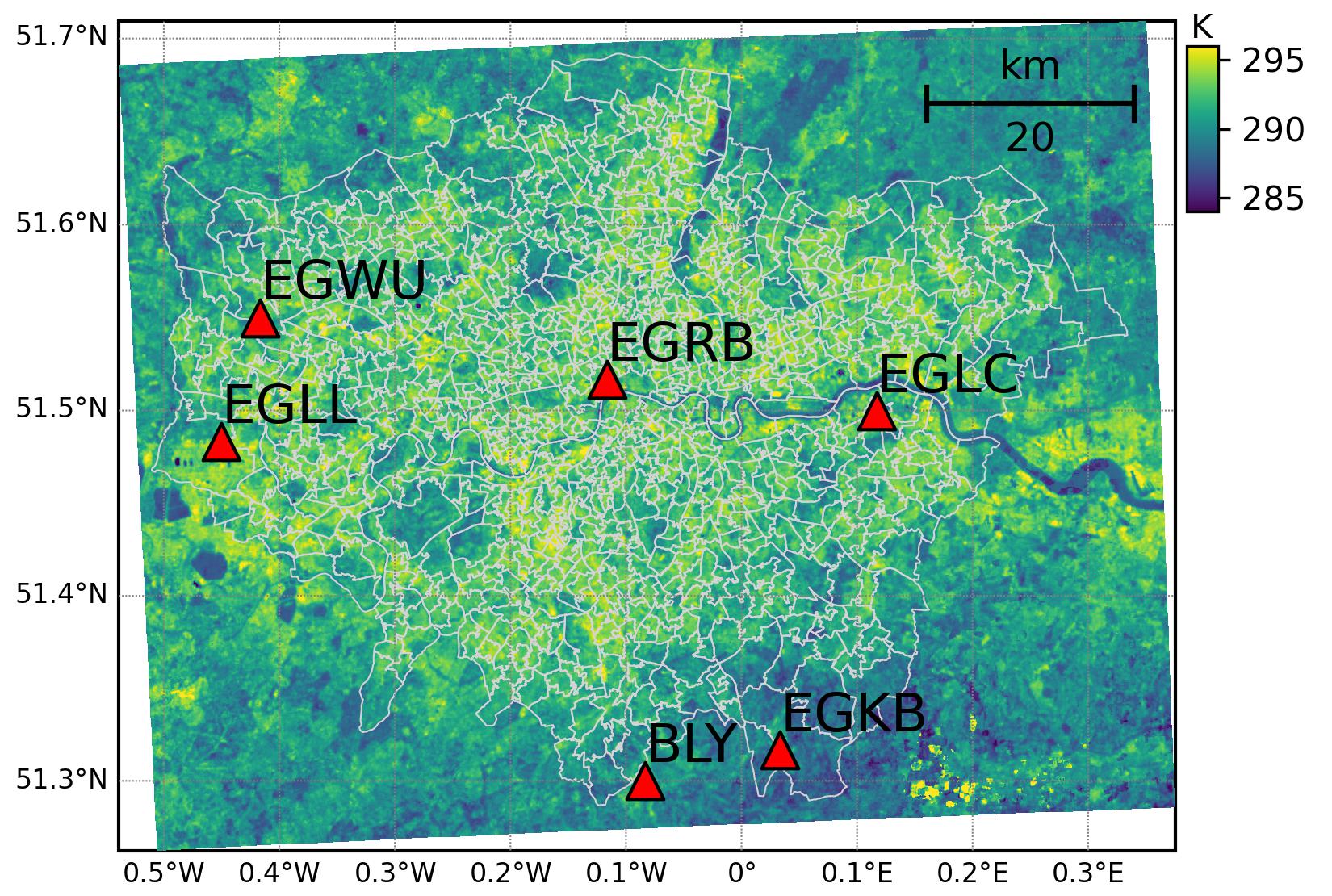}
    \includegraphics[width=0.9\linewidth]{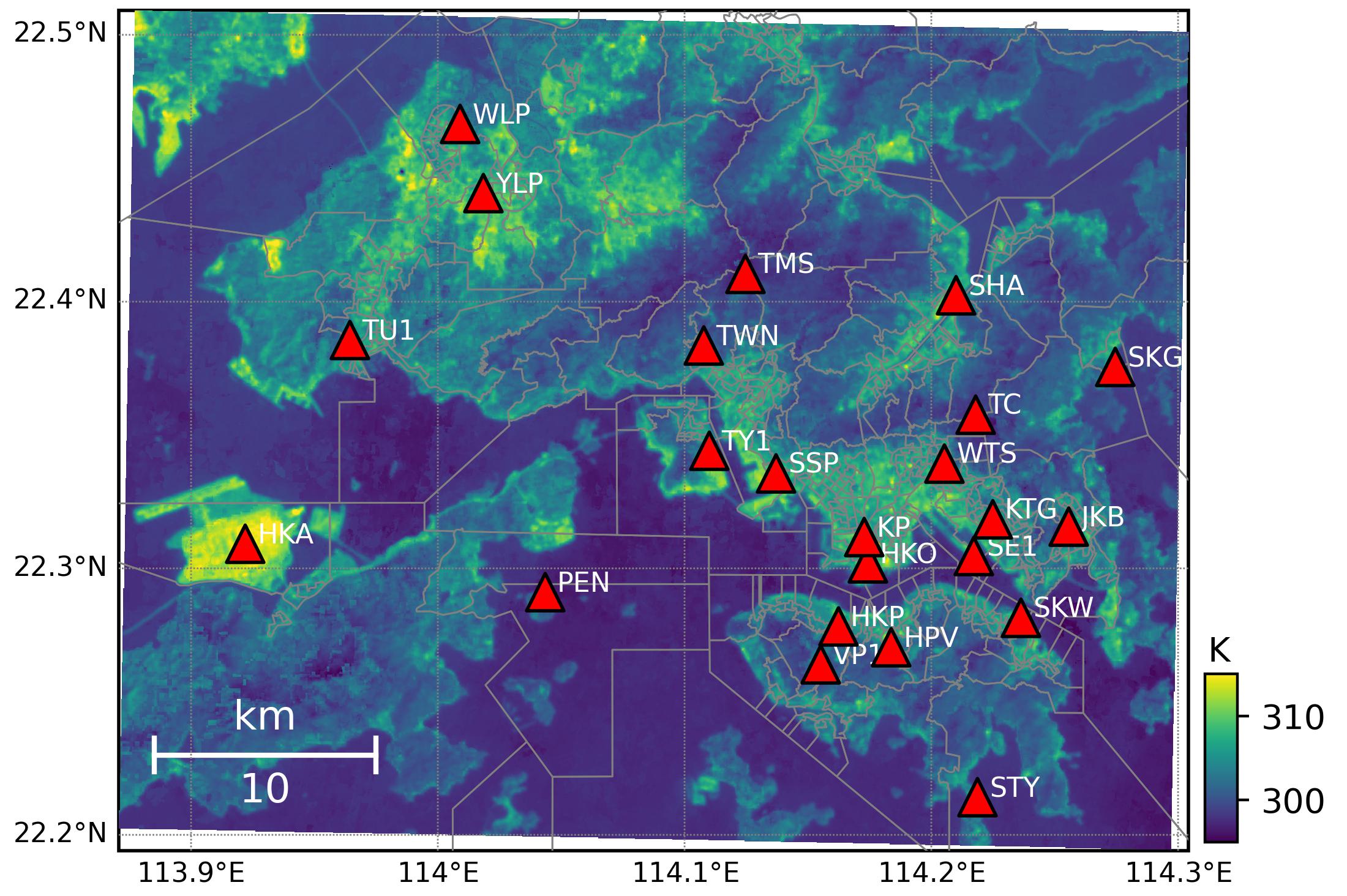}
    \caption{Meteorological stations of the three cities, overlapped with annual mean temperature and geographical unit boundaries. NYC: 7 stations over 135 merged zip codes; London: 6 stations over 983 Middle Super Output Area; Hong Kong: 23 stations over 452 District Council Ordinary Election Constituency Boundaries.}
    \label{fig:AirTempStations}
\end{figure}

\subsection{Data and Preprocessing}
\subsubsection{Landsat Land Surface Temperature}
We obtained all 2023 Landsat 8 and 9 Collection 2 Level-2 science products and used the analysis-ready LST data from the thermal band (Band 10). NYC is completely located within two Landsat scenes (Path 013 Row 032, Path 014 Row 032), with a total number of 92 scenes. London is completely covered by one scene (Path 201 Row 024), and has its majority covered by another (Path 202 Row 024), with a total number of 86 scenes. Hong Kong has most of its urban areas covered by two scenes (Path 121 Row 045, Path 122 Row 044), with a total number of 91 scenes. 

Before conducting the experiments, we went through a series of preprocessing, mainly on cloud masking and removal. We first checked the cloud band from the Level-2 products and masked out all data labeled as unclear. The cloud masks are generated from the Fmask algorithm and are not 100\% perfect as noted from data description. We then double-checked visually with the imagery and cloud masks using a 3$\times$3 split. Within each split, if the pixels were mostly flagged as clear but were in fact covered by clouds, we masked out all the pixels within the split. In some cases, within each split, all pixels within the split appeared clear and most pixels were flagged as clear, but some pixels were flagged as not clear---mostly at coastline---we flagged this split as clear. After preprocessing, we have three types of data: dates under clear-sky situations, dates under partially-cloudy situations, and dates without valid observations (100\% cloudy or no Landsat overpass).

\subsubsection{ERA5 Reanalysis LST Data}
The proposed method DELAG includes an enhanced 4-parameter ATC model that has the capability to capture daily temperature fluctuation. The additional variable in the enhanced ATC model needs to be a continuous temperature record, and we used the skin temperature from ERA5 daily aggregated data~\cite{hersbach2020era5}. The ERA5 reanalysis products have a resolution of 11~km. Each pixel is matched to the nearest ERA5 LST value. 

\subsubsection{Annual Mean Spectral Reflectance}
In order to link valid-observed pixels to cloud-covered pixels, the proposed method DELAG uses Gaussian processes to model the residuals between observed LST and ATC-reconstructed LST. The residuals are related to ground objects, which can be distinguished through spectral reflectance. We obtained the mean surface reflectance from 2022-2024 for four spectral bands (red, green, blue, and near infrared) from the Sentinel-2 products. The data is then resampled and matched with the Landsat ones. We obtained through Sentinel-2 instead of Landsat because Sentinel-2 has a higher revisit frequency, but the influence is negligible as seen from a sensitivity analysis. We also added the x and y coordinates, along with the spectral bands, as the features used in Gaussian processes.

\subsubsection{Near-surface Air Temperature Data}
There is no in-situ LST site within the three cities, thus a direct comparison with in-situ LST in our study areas is not possible. However, given that most real-world applications require near-surface air temperature instead of LST, and many seamless LST data will be used for generating near-surface air temperature, we can indirectly validate the reconstruction with near-surface air temperature data (details in method Section~\ref{sssec:ValidationAirTemp}). This is an approach commonly used in the existing literature~\cite{yang2024annual, zhao2020reconstruction}. For this purpose, we obtained air temperature data from meteorological stations. For New York City, we obtained data from 7 stations from the Automated Surface/Weather Observing Systems (ASOS/AWOS) maintained by the National Weather Service. For London, there are 6 meteorological stations. These data were obtained through Meteostat, an open source Python library that archives meteorological data from public sources. For Hong Kong, we obtained air temperature data from 23 stations from the Hong Kong Observatory. The locations of the meteorological stations in the three cities can be found in Fig.~\ref{fig:AirTempStations}.

\section{Methodology}
\label{sec:DataMethod}

\subsection{DELAG: Deep Ensemble Learning With Annual Temperature Cycles and Gaussian Processes}
\label{sec:DELAG}
The proposed method DELAG is a two-stage model, including enhanced annual temperature cycles (ATC) and Gaussian processes (GP). In the first stage, for each pixel, DELAG reconstructs LST using valid-observed data through an enhanced ATC model. The model solution is solved using deep ensemble learning by minimizing the loss function via Adam optimization. A total of 200 model snapshots are obtained to generate ensemble predictions and quantify uncertainty. In the second stage, the residuals between ATC-reconstructed LST and the valid-observed Landsat LST from the non cloud-covered pixels are calculated. Within each day, the residuals are related to certain surface properties, which can be used as features to predict the residuals. The residuals, as a surface of random processes, can be modeled through Gaussian processes based on valid-observed pixels. Within the same day, this surface can be generalized to cloud-covered pixels, thus also achieving LST reconstruction for cover-covered pixels on dates with valid-observed LST. The uncertainty on dates with valid-observed LST is then quantified from both models from the first stage (ATC) and second stage (GP). For dates without any valid-observed LST (100\% cloud cover or no Landsat overpass), only the first part of the estimations will be applied. 

Overall, the proposed method DELAG can be expressed as  
\begin{equation}
    \label{eq:model1}
    \mathcal{T}_{s}^{R} \left(d\right) = \mathcal{T}_{s}^{ATC}\left(d\right) + \mathcal{T}_{s}^{GP} \left(d\right) \;,
\end{equation}
where $d$ is day of year, $s=[x,y]$ is individual pixel's indexes,  $\mathcal{T}_{s}^{ATC}\left(d \right)$ is the ATC-reconstructed temperature as a function of day of year of pixel $s$, and $\mathcal{T}_{s}^{GP}\left(d \right)$ is the temperature residual estimated from GP. 

On day of year $d$, given the coarse-resolution ERA5 reanalysis LST value $\mathcal{T}^{c} \left(d \right)$, a set of valid-observed Landsat pixels $\bf{s}_{\mathit{d}}$, and the corresponding valid-observed LST data $\mathcal{T} \left(d, \bf{s}_{\mathit{d} } \right)$, Equation \ref{eq:model1} can be written out as 
\begin{equation}
\begin{split}
    \mathcal{T}_{s}^{R} \left(d \right)  = &  C_{s} + A_s \cos \left( 2\pi / 365 \left(d - \phi_s \right) \right) + b_s \mathcal{T}^c \left( d \right)  \\ 
   & + \mathcal{GP}_d \left(s, \bf{s}_{\it{d}} \right)  \;,
\end{split}
\end{equation}
where $C_{s} + A_s \cos \left(2\pi / 365 \left(d - \phi_s\right) \right)$ is the classic ATC model with $\{C_{s},A_s,\phi_s\}$ as three location-dependent ATC parameters, $b_s$ is a location-specific parameter to capture daily fluctuation (making it an enhanced ATC). And $\mathcal{GP}_d \left(s, \bf{s}_{\it{d}} \right)$ is to use a daily-specific GP model $\mathcal{GP}_d \left(\cdot \right)$ to account for the temperature residuals between valid-observed Landsat LST data and ATC-reconstructed LST by modeling the relationship between certain surface properties and the valid-observed LST residuals $\mathcal{T} (d, \bf{s}_{\mathit{d} }) = \{ \mathcal{T}_{\mathit{k},\mathit{d}} \}_{\mathit{k} \in \bf{s}_\mathit{d}  } $ from a set of valid-observed pixels $\bf{s}_{\it{d}}$.

\subsubsection{Enhanced Annual Temperature Cycles}
We use the PyTorch deep learning framework to find the solutions of the model. In the first stage (eATC), we minimize the L1 loss between the observed LST $\mathcal{T}$ and the eATC-reconstructed LST $\hat{\mathcal{T}}^{ATC}$. This process is equivalent to maximizing a Laplace likelihood, and can be written out as (with the location index $s$ omitted)
\begin{equation}
    \label{eq:l1loss}
    p \left(\mathcal{T}  \mid \omega \right) \propto \exp \left(\frac{ - | \mathcal{T} - \hat{\mathcal{T}}^{ATC}  |  }{\beta}   \right)   \; ,
\end{equation}
where $\omega = \{C, A, \phi, b \}$ is the parameters of eATC that is used to calculate $\hat{\mathcal{T}}^{ATC}$, and $\beta$ is a scale factor of the Laplace distribution.

\subsubsection{Gaussian Processes to Model LST Residuals}
On the same date, LST residuals ($ \mathcal{T}^{\epsilon} = \mathcal{T} - \mathcal{T}^{ATC}$) are related to certain surface property. We can then model this surface from GP by linking certain features and LST residuals from the valid-observed pixels. This GP model can be used to predict the LST residuals to the cloud-covered pixels, as we assume the LST residuals follow the same random processes related to surface property within the same day. For features, we use the annual mean spectral reflectance from red, green, blue, and near infrared bands from Sentinel-2, and add the two coordinate variables to capture some spatial trends. The GP model is built with GPyTorch, a GP framework built on PyTorch for scalable Gaussian processes through variational and Blackbox Matrix-Matrix inference~\cite{gardner2018gpytorch}. 

On day of year $d$, we assume the LST residuals $\mathcal{T}^{\epsilon}$ follow a GP with mean function $m \left(s \right)$ and covariance function $k \left(\cdot, \cdot \right)$ (with $d$ omitted): 
\begin{equation}
\label{eq:gp}
\mathcal{T}^{\epsilon} \left( s \right) \sim \GP \left(m \left( s \right), k \left(\cdot, \cdot \right) \right)\;,
\end{equation}
where $k \left(s, s' \right)$ defines the covariance between two pixels $\left( s, s' \right)$ over LST residuals $\mathcal{T}^{\epsilon} \left( s \right)$ and $\mathcal{T}^{\epsilon} \left( s' \right)$ based on their similarity. 

Given known LST residuals $\mathcal{T}^{\epsilon} \left( \mathbf{s}_{\mathit{d} } \right) $ from valid-observed Landsat LST pixels $\mathbf{s}_\mathit{d} $, we have their feature representation (spectral bands) $\mathbf{X}_{ \mathbf{s}_\mathit{d}  } $ and a covariance matrix $\mathbf{K}_{\mathbf{s}_\mathit{d} } $ that records the covariance between all valid-observed Landsat LST (training data points). We train the model by maximizing the log-likelihood function, which can be written as (with $\bf{s}_\mathit{d}$ omitted): 
\begin{align}
  &\begin{aligned}
  & \log p \left( \mathcal{T}^{\epsilon}  \mid \mathbf{X}, \mathbf{K}, \Theta \right) \\ & =   - \frac{1}{2} \left( \mathcal{T}^{\epsilon} -m \left( \mathbf{X} \right)  \right)^\top \mathbf{K}^{-1} \left( \mathcal{T}^{\epsilon} - m \left( \mathbf{X} \right) \right)  \\
  & \;\;\;\;\;\;\;  - \frac{1}{2} \log |\mathbf{K}| - \frac{n}{2} \log \left( 2\pi \right)   \;,
  \end{aligned}
\end{align}
where $\Theta$ is the hyperparameters within the covariance function; in our experiments, we used the radial basis function (RBF) kernel. 

\subsubsection{LST Residuals for Cloud-Covered Pixels}
Once the GP model is built, we can estimate the LST residuals over cloud-cover pixels. On day of year $d$, given non-cloud-covered pixels $\mathbf{s}_{\mathit{d} }$ and cloud-cover pixels $\mathbf{s}^{*}_{\mathit{d}}$, their feature representations $\mathbf{x}_{\mathbf{s}}$ and $\mathbf{x}_{\mathbf{s}^{*}}$, the mean of the LST residuals modeled by GP ($\mathcal{T}^{GP}_{\mathbf{s}^{*}}$) for the cloud-cover pixels is:
\begin{equation}
     \mathcal{T}^{GP}_{\mathbf{s}^{*}} = \mu\left(\mathbf{x}_{\mathbf{s}^{*}}\right) = \mathbf{k}_{\mathbf{s}^*}^\top (\mathbf{K} + \sigma^2 \mathbf{I})^{-1} \mathcal{T}^{\epsilon}_{\mathbf{s}} \; ,
\end{equation}
where $\mathbf{k}_{\mathbf{s}^*} = [k \left(\mathbf{x}_i, \mathbf{x}_{\mathbf{s}^*} \right)]_{i=1}^n$ is the covariance vector between training points and $\mathbf{x}_{s^*}$, $\mathbf{K}$ is the $n \times n$ covariance matrix of the training points, $\sigma^2$ is the observed noise. The prediction variance, which is used for uncertainty quantification, is calculated as
\begin{equation}
    \sigma^2(\mathbf{x}_{\mathbf{s}^{*}}) = k(\mathbf{x}_{\mathbf{s}^{*}}, \mathbf{x}_{\mathbf{s}^{*}}) - \mathbf{k}_{\mathbf{s}^*}^\top (\mathbf{K} + \sigma^2 \mathbf{I})^{-1} \mathbf{k}_{\mathbf{s}^*} \; .
\end{equation}
The calculation is accelerated through GPyTorch.

\subsubsection{Deep Ensemble Learning to Find Model Solutions and Achieve Uncertainty Quantification}
\begin{figure}[!t]
    \centering
    \includegraphics[width=0.9\linewidth]{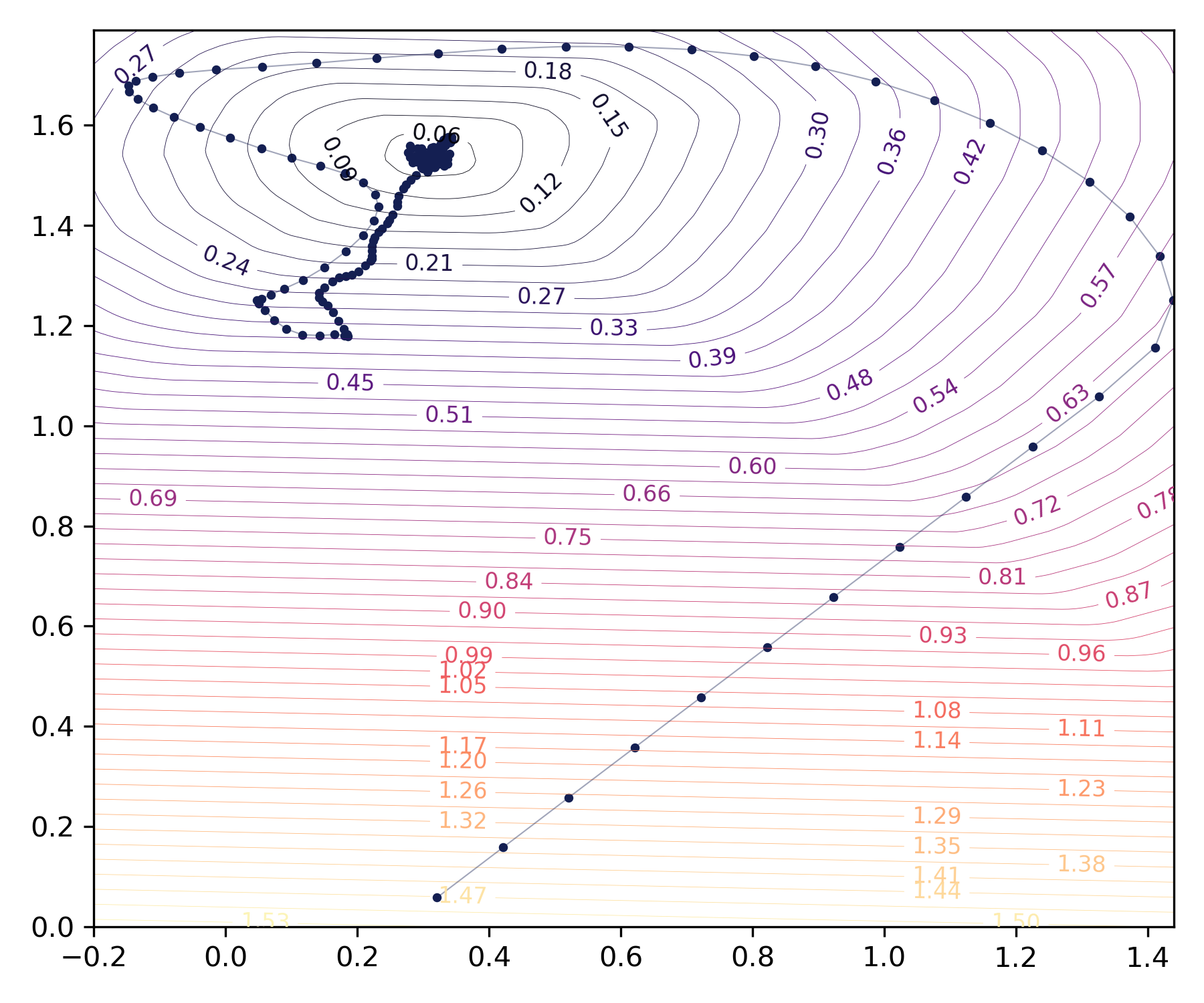}
    \caption{Demo showing the loss surface and model trajectory during training. The x- and y-axes represent unitless, compressed parameter dimensions. As training progresses, the loss decreases substantially, with 334 out of 425 snapshots falling below the lowest contour level (0.06). DELAG leverages these diverse solutions as an ensemble to produce robust predictions with uncertainty quantification.}
    \label{fig:LossSurface}
\end{figure}

Given the complexity of the model as a non-convex function, it is very difficult, if not impossible, to find the global optimized set of parameters. In the PyTorch deep learning framework, we use the Adam optimizer (a variant of stochastic gradient descent)~\cite{kingma2014adam} to minimize the L1 loss to find the optimized solutions in the first step. In training the model, the training loss will eventually flatten and fluctuate, but cannot reach a global minimum. In Fig.~\ref{fig:LossSurface}, near the end of training, 334 out of 425 snapshots all fall below the smallest contour level (loss$<$0.06). At this phase of the training, all snapshots are acceptable optimized solutions of the model. With each iteration, the parameters update again, resulting another snapshot of model parameters. The disagreement drawn from these snapshots of parameters is large enough, and utilizing the disagreements among these snapshot solutions has become a popular method to achieve ensemble learning or quantify uncertainty in deep learning~\cite{huang2017snapshot, izmailov2018averaging, garipov2018loss}. Ensemble learning to achieve more robust and reliable prediction is also common in weather predictions and remote sensing~\cite{gneiting2005weather, liu2020active}. 

We take advantage of these snapshot solutions to generate ensemble predictions, revising Equation \ref{eq:l1loss} to include an index $j$ for these optimized solutions: 
\begin{equation}
    p \left(\mathcal{T}  \mid \omega \right)  =  \frac{1}{J} \sum_j p \left(\mathcal{T}  \mid \omega_j \right) \;.
\end{equation}
This ensemble process can be seen as the posterior predictive distribution over a uniform prior. 

\subsubsection{Uncertainty Quantification}
The variance of Landsat LST comes from the cross-day variance (modeled via enhanced ATC) and within-day variance (modeled via GP). Using the law of total variance, we decompose the total variance of $\mathcal{T}$ into two components:
\begin{equation}
\text{Var} \left( \mathcal{T} \right) = \mathbb{E} \left[ \text{Var} \left( \mathcal{T}  \mid D \right) \right] + \text{Var} \left( \mathbb{E} \left[ \mathcal{T}  \mid D \right] \right) \;,
\end{equation}
where $\mathbb{E} \left[ \text{Var} \left( \mathcal{T} | D \right) \right]$ is the within-day variance and $\text{Var} \left( \mathbb{E} \left[ \mathcal{T} | D \right] \right)$ is the cross-day variance. Therefore, we can simply add the quantified uncertainty together to achieve the final uncertainty quantification: 
\begin{equation}
\text{Var} \left( \mathcal{T} \right) = \text{Var} \left( \mathcal{T}^{ATC} \right) + \text{Var} \left( \mathcal{T}^{GP} \right) \; .
\end{equation}

In DELAG, we obtain 200 snapshots (200 predictions) from the enhanced ATC models. We can then easily obtain the 95\% confidence intervals (CI) from the 200 snapshots. GP models inherently generate uncertainty quantification and prediction intervals. We then add the upper and lower predictions from the two stages to estimate the 95\% CI. The uncertainty quantification applies to dates with clear-sky situations and dates with partially-cloudy situations. For dates without any valid-observed LST, the uncertainty quantification reduces to enhanced ATC only.

\subsection{Validation Strategies}
\label{ssec:SalidationStrategies}
Considering the four different situations (clear sky, partially cloudy, no valid Landsat pixels due to 100\% cloud cover, and no valid Landsat pixels due to no Landsat overpass), we design three different validation strategies. 

\subsubsection{Validation With Landsat Clear-Sky Observations Under Real-World Cloud Patterns}
We first validate the result using clear-sky observations under real-world cloud patterns. We randomly select a few dates that have most clear pixels and then mask out some pixels as test data (under some hypothetical cloud pattern). These hypothetical masks are randomly selected from some real-world cloud patterns observed from other dates. This validation strategy can validate results under clear-sky situations.

\subsubsection{Validation With Partially-Cloudy Data}
For dates with partially-cloudy situations, the percentage of clouds has a significant impact on the solar energy balance. Under partially-cloudy situations, the valid-observed LST pixels should not be considered as clear-sky pixels. If the percentage of clouds is high enough, the valid-observed LST pixels are close to the overcast situations, as the valid-observed pixels are blocked by clouds and non-observable most of the time (depending on percentage of clouds). Therefore, we separately validate the results under partially-cloudy situations, focusing on dates with more than 80\% cloud covers. For this setting, we hold out 20\% of the valid-observed pixels and use them as test data. 

\subsubsection{Validation Via Estimating Near-Surface Air Temperature}
\label{sssec:ValidationAirTemp}
Since there is no in-situ LST site in the three cities, a direct comparison with in-situ LST data is not possible. However, because there is a strong correlation between LST and near-surface air temperature, especially under cloudy conditions~\cite{vancutsem2010evaluation}, we can indirectly validate the results with in-situ air temperature data, an approach commonly used in the existing literature~\cite{yang2024annual,zhao2020reconstruction}. More importantly, the capability to generate accurate near-surface air temperature data is critical for many applications, especially for those related to health sciences~\cite{white2013validating}. If the reconstructed LST can achieve comparable results in estimating near-surface air temperature compared with the valid-observed LST, the reconstruction is considered successful. For this purpose, similar to previous studies, we adopt a simple statistical approach developed by Janatian~\emph{et al.}~(2017)~\cite{janatian2017statistical} to generate near-surface air temperature based on LST: 
\begin{equation}
    \mathcal{T}_{a} = \mathcal{T}_{lst} \times \alpha_1 + V_{ndvi} \times \alpha_2  + V_{elv} \times \alpha_3 + V_{sol} \times \alpha_4 + V_{sza} \times \alpha_5 + c \;,
\end{equation}
This is a linear equation that consists of five variables. Apart from LST, the equation includes NDVI ($V_{ndvi}$), elevation ($V_{elv}$) and solar radiation factor ($V_{sol}$) to account for local conditions, and solar zenith angle ($V_{sza}$) to account for seasonal changes. We apply this equation once for all meteorological stations within a city, but separately for the valid-observed LST and the reconstructed LST. The LST reconstruction results are considered good when the generation of air temperature from reconstructed LST has similar performance from the clear-sky LST. 

\subsubsection{Evaluation Metrics}
We use four standard metrics for validation: mean absolute error (MAE), root mean square error (RMSE), coefficient of determination (R$^2$), and bias (Bias).

\subsection{Training Setup}
The experiments were conducted on Python using the PyTorch framework~\cite{paszke2019pytorch}. For ATC fitting, we used the Adam optimizer with a learning rate of 0.1 for 1200 epochs. The 200 ensemble predictions were obtained through every 4 epochs in the final 800 epochs. For Gaussian processes, we used GPyTorch~\cite{gardner2018gpytorch}, a popular GP framework built upon PyTorch. We used a mini batch of 1024 and the variational inference with 512 inducing points for fast GP computation. The training was conducted with the Adam optimizer with a learning rate of 0.05 for 50 epochs and then of 0.005 for 10 epochs. The training settings apply to all experiments.

\subsection{Experimental Setup}
We conducted three sets of experiments to examine the performance of the proposed method under different cloud conditions. In the first set of experiments, we subset a part of the core urban areas for testing, identified a few dates with 0\% cloud cover, and randomly selected real-world cloud patterns from a different date as the hypothetical clouds. This set of experiments is to test the proposed method in reconstructing LST under data-rich, clear-sky situations. Test data are from clear-sky LST pixels that are masked out with some real-world cloud pattern from other dates during training. The second set of experiments was conducted over all data, with this validation strategy specifically targeting heavily cloudy situations ($>$80\%). On these dates, the valid-observed LST is not clear-sky LST but rather similar to overcast situations because of the high percentage of clouds (the satellite was only able to record data because there were a few gaps between the clouds). Test data are from 20\% of the valid-observed LST under the heavily-cloudy situations that were held out during training. This set of experiments is to test the proposed method in partially-cloudy and heavily-cloudy situations. In the third set of experiments, as there is no in-situ LST site within the three cities, we are not able to test the proposed method under completely overcast (100\%) or without Landsat overpass situations. Alternatively, we adopt a common approach in the literature~\cite{yang2024annual, zhao2020reconstruction} and validate the reconstruction results indirectly by comparing its capability to estimate near-surface air temperature data (detailed earlier in section~\ref{sssec:ValidationAirTemp}). This set of experiments can test LST reconstruction under all situations, but we focus on cloudy situations or without Landsat overpass. These above-mentioned three validation strategies are detailed in the section~\ref{ssec:SalidationStrategies}.

\section{Results and Analysis}
\label{sec:result}
\subsection{Integrating ERA5 Reanalysis Data and Landsat Observations for LST Reconstruction}
\begin{figure*}[!t]
    \centering
    \includegraphics[width=0.98\linewidth]{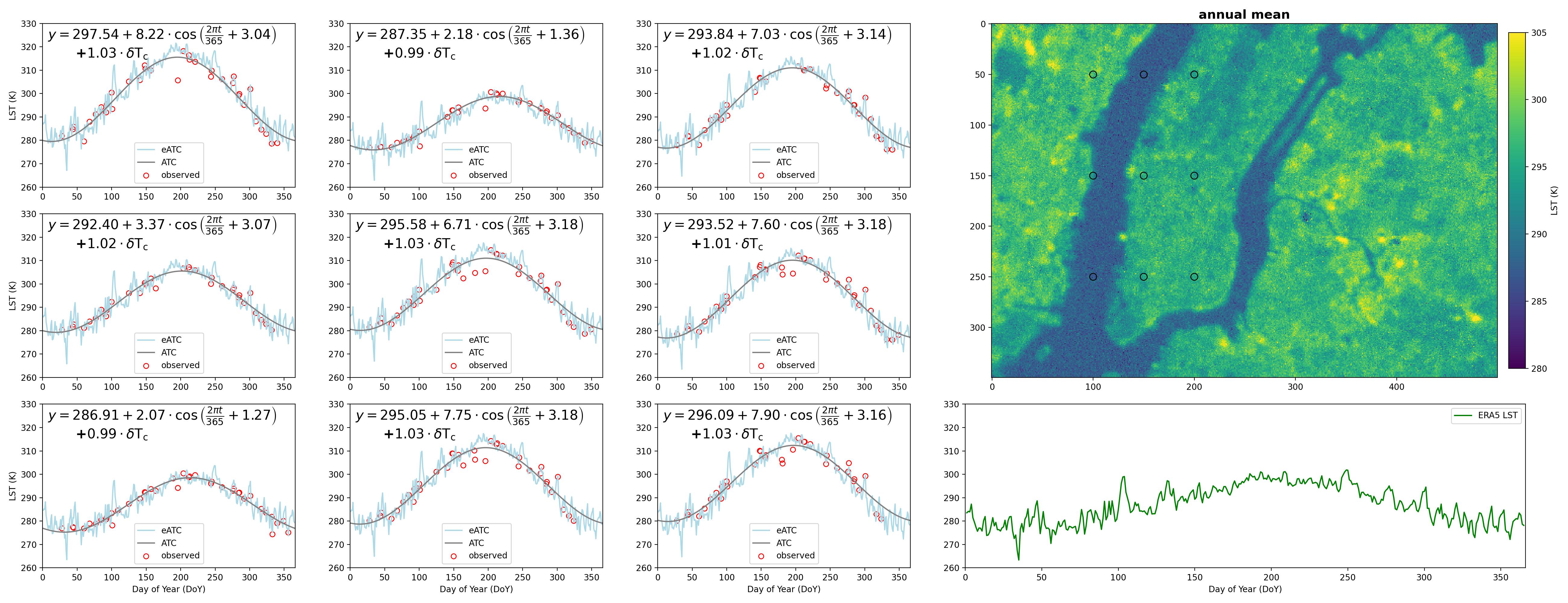}
    \caption{Demo of fitted eATC models for 9 Landsat pixels within a single ERA5 data pixel. The Landsat pixels, marked with circles in the image, are arranged in the subplots to reflect their relative positions.}
    \label{fig:fittedmodels}
\end{figure*}

We first demonstrate that, even within the same ERA5 reanalysis data pixel (11~km), the reconstructed LST can vary substantially due to differences in Landsat observations and the resulting fitted parameters in the eATC models. As shown in Fig.~\ref{fig:fittedmodels}, the Landsat pixels, which all correspond to the same ERA5 data pixel, have a wide range of temperature values. The pixel locations are marked in circle in the image, and their relative positions are the same as arranged in the plots. Through different fitted parameters, the eATC models successfully captured the majority of the temporal information. Specifically, the first parameter, representing the annual mean temperature, ranges from 286.91 to 297.54~K. The water pixels (1st row 2nd column, 3rd row 1st column) have lower annual mean temperature, as expected. The water pixels also have a smaller amplitude (2nd parameter), being 2.07 and 2.18~K, compared to land and urban pixels' 6.72-8.22~K. (The 2nd row 1st column is a land-water boundary pixel with a 3.37~K amplitude.) The linear term on the ERA5 reanalysis data further captures the daily fluctuation through varying the 4th parameter, ranging from 0.99 to 1.03. As the ERA5 data pixel can be considered as the mean of a larger area, its trend is therefore a signal combining all individual Landsat pixels falling within the ERA5 data pixel. Assuming a linear relationship between individual Landsat pixels and the ERA5 data can partially recover Landsat LST data, and, combined with the original three parameters of ATC, from the demo here we can see that the majority of the information is recovered through the eATC model (Fig.~\ref{fig:fittedmodels}). The remaining residuals are then recovered through Gaussian processes.

\subsection{Residual Surface Analysis Before Gaussian Processes}
\begin{figure*}[htbp]
    \centering
    \includegraphics[width=0.98\linewidth]{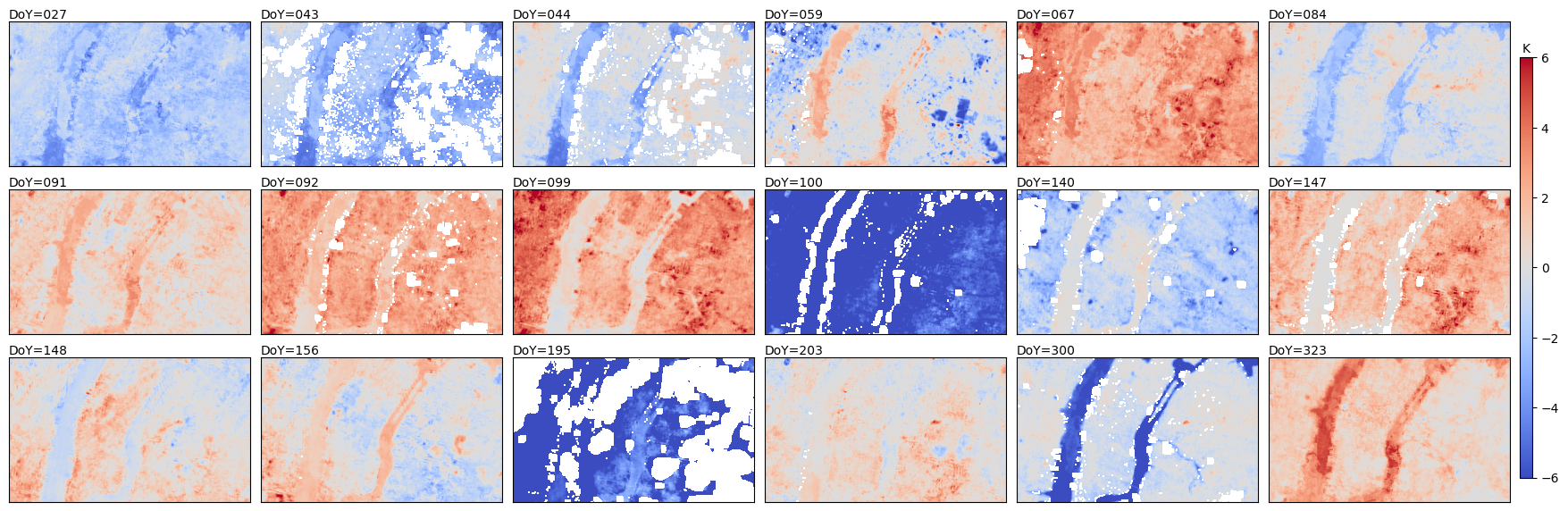}
    \caption{Various patterns of the residual surfaces from different days.}
    \label{fig:ResidualSurfaces}
\end{figure*}

If residual surfaces of individual days are related to each other, we can then generalize residual surfaces from days with observations to days without observations. Unfortunately, the residual surface is not identical each day, as the information captured through the eATC model is unbiased and scattered around the observations, meaning that the eATC reconstructed LST can be higher or lower than the actual value. We visualize the residual surfaces from 18 selected days of the NYC data in Fig.~\ref{fig:ResidualSurfaces}, showing different patterns across different days. Nonetheless, on the same day, the residual surface is related to ground surface materials (a notable patten is the residuals over water and land areas), which we can use Gaussian processes to model. 

\subsection{Results Under Clear-Sky Situations With Real-World Cloud Patterns}
\begin{figure*}[htbp]
    \centering
    \includegraphics[width=0.3\linewidth]{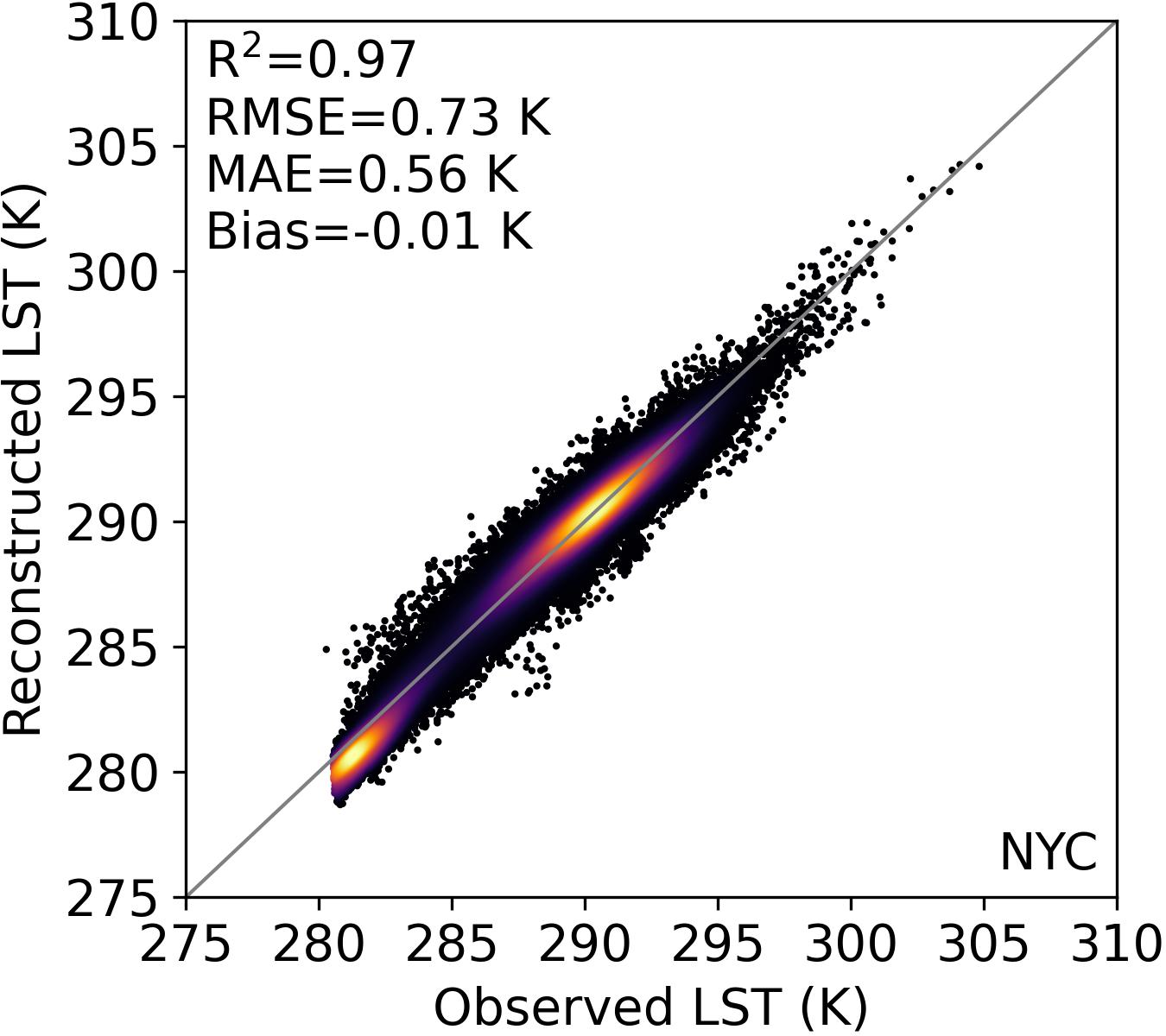}
    \includegraphics[width=0.3\linewidth]{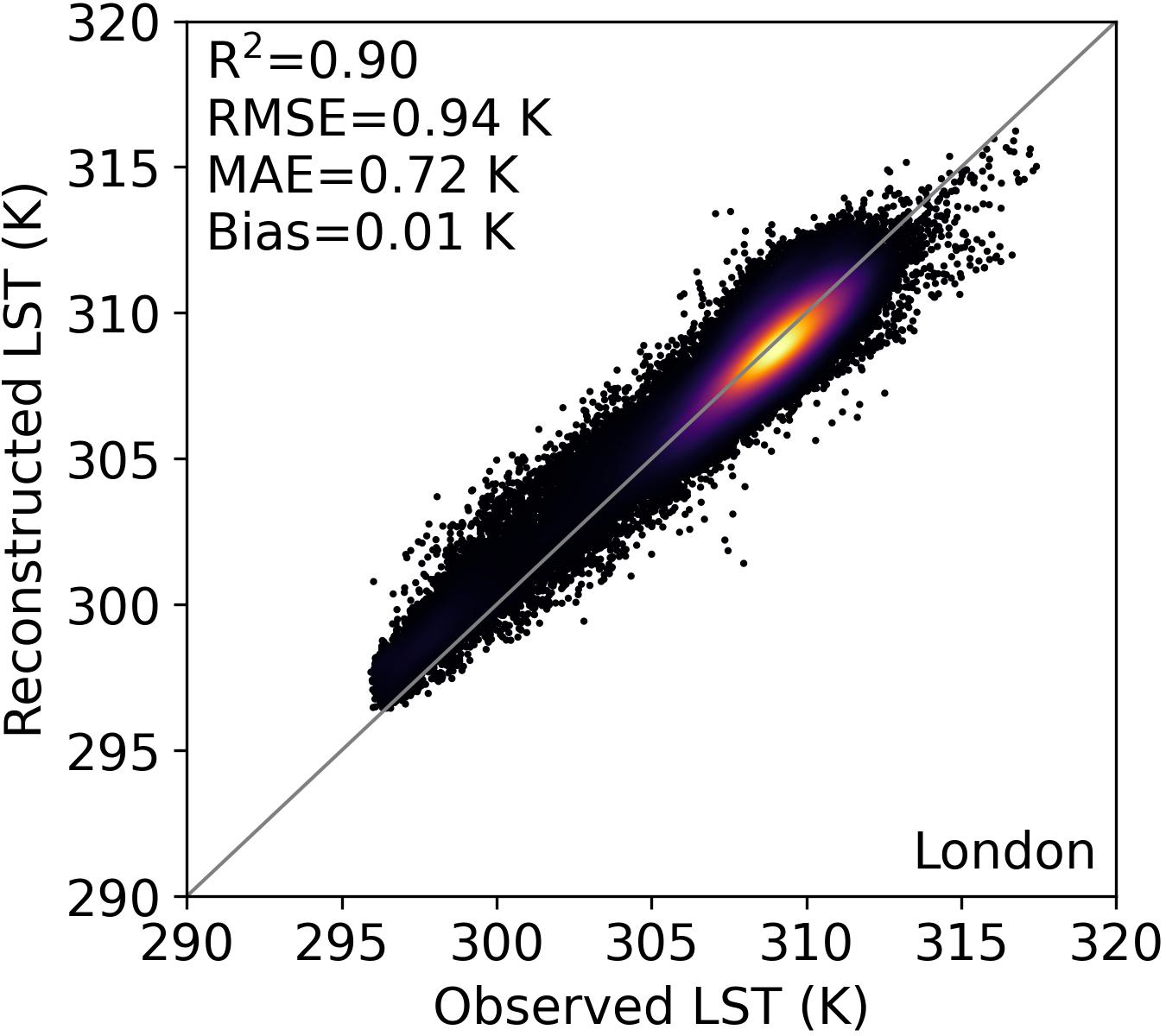}
    \includegraphics[width=0.3\linewidth]{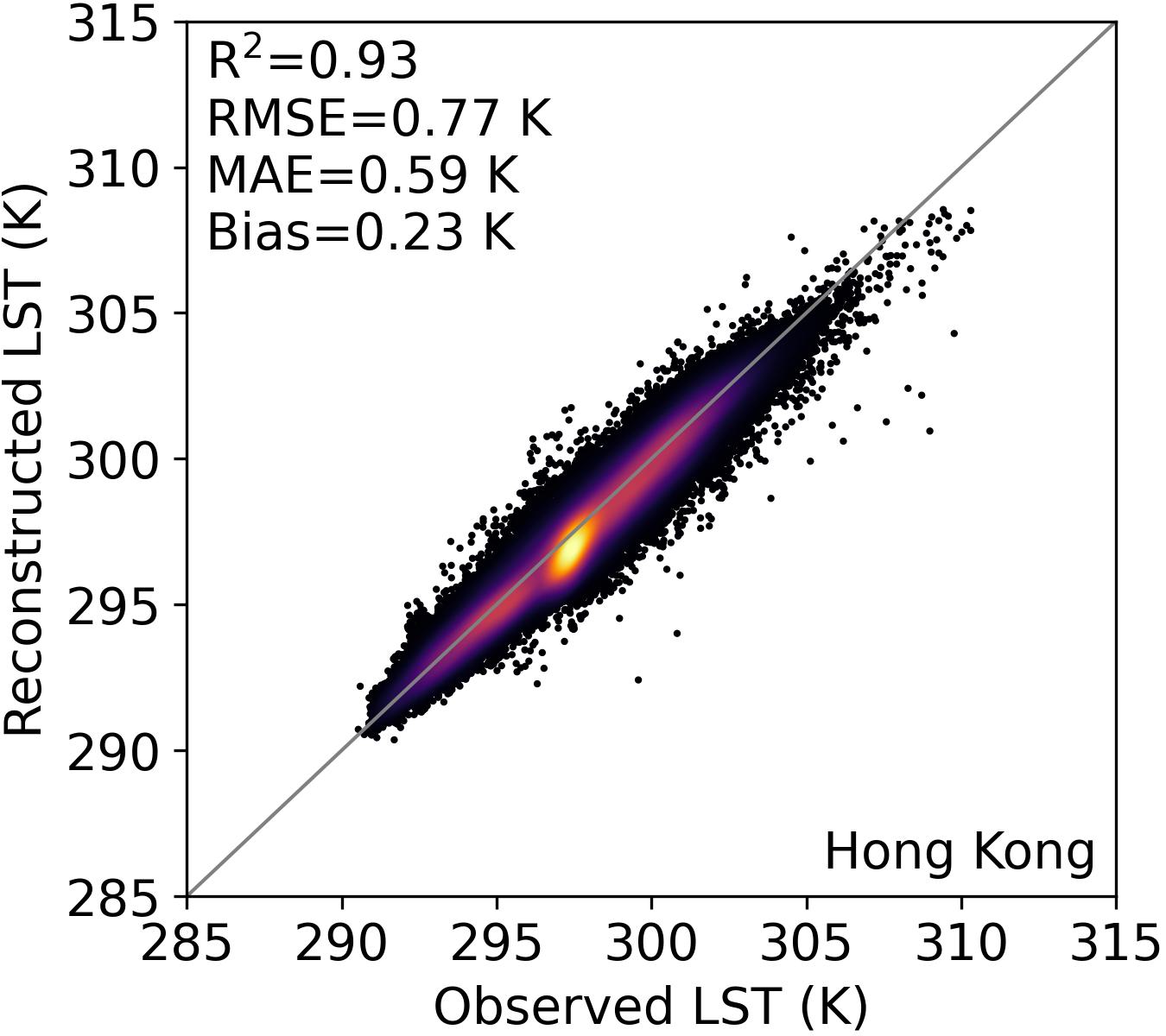}
    \caption{Scatter plot of reconstructed versus observed LST under clear-sky conditions, using real-world cloud patterns from other days.}
    \label{fig:PlotClearSkyRealCloud}
\end{figure*}

\begin{figure*}[!t]
   \centering
    \includegraphics[width=0.92\linewidth]{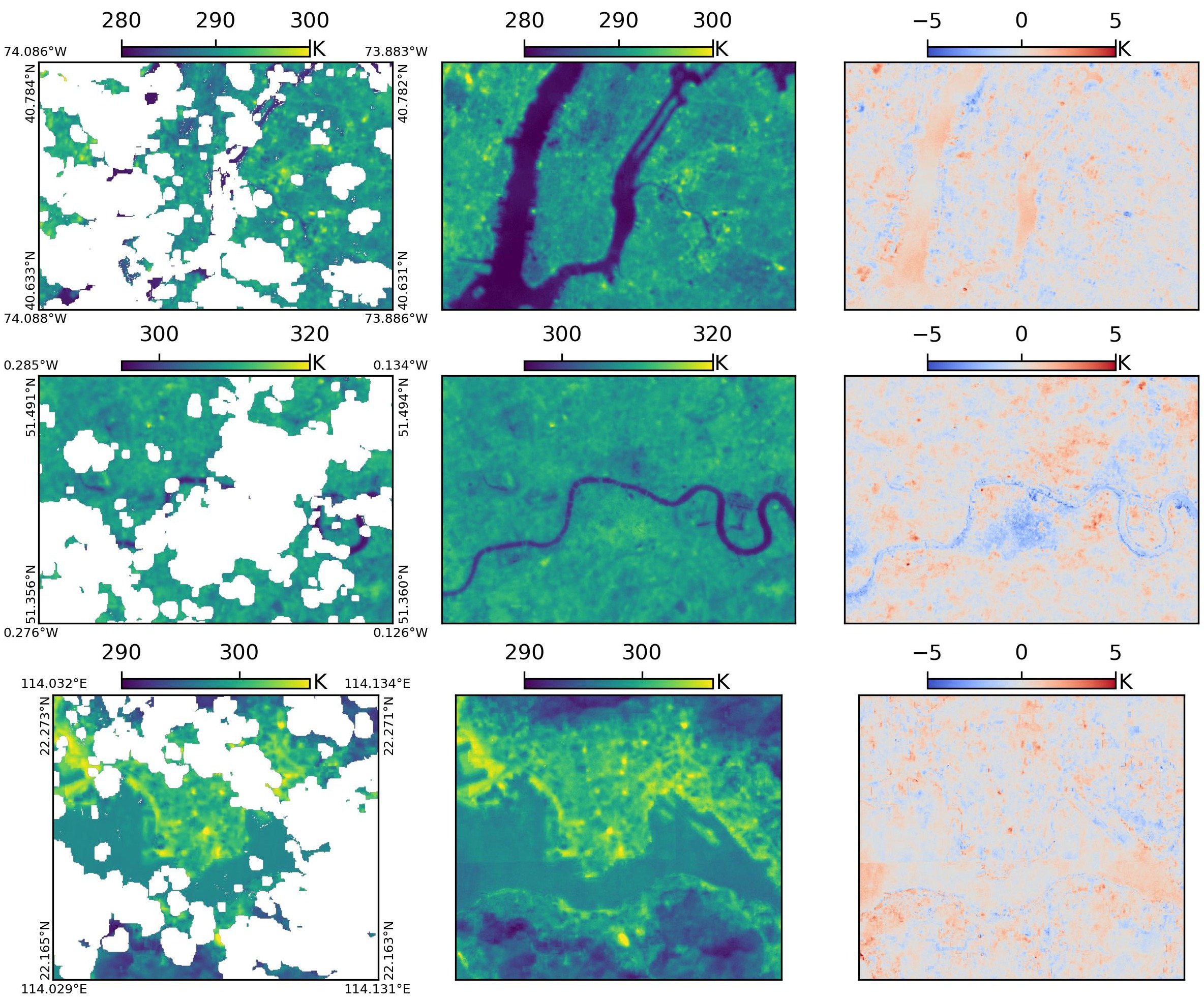}
    \caption{Results under clear-sky conditions using real-world cloud patterns from other days. Columns display: Observed LST (with real-world cloud patterns), Reconstructed LST, and Residuals. Rows represent NYC, London, and Hong Kong.}
    \label{fig:ImageClearSkyRealCloud}
\end{figure*}

The results under clear-sky situations with real-world hypothetical cloud patterns from another day are shown in Fig.~\ref{fig:PlotClearSkyRealCloud} with scatter plots, and in Fig.~\ref{fig:ImageClearSkyRealCloud} with the reconstruction visualizations including the cloud patterns, reconstructed LST, and residuals. The proposed method successfully achieved LST reconstruction under clear-sky situations with real-world cloud patterns from other dates. Under 52.0\% hypothetical cloud coverage, the proposed method reached a reconstruction result of R$^2$=0.97, RMSE=0.73 K, MAE=0.56 K, and Bias=-0.01 K for NYC (prediction date: 2023-04-01, cloud pattern date: 2023-06-04). Under 52.03\% hypothetical cloud cover, the proposed method reached a reconstruction result of R$^2$=0.90, RMSE=0.96 K, MAE=0.74 K, and Bias=-0.27 K for London (prediction date: 2023-09-06, cloud pattern date: 2023-06-18). Under 56.2\% hypothetical cloud coverage, the proposed method reached a reconstruction result of R$^2$=0.93, RMSE=0.74 K, MAE=0.56 K, and Bias=-0.02 K for Hong Kong (prediction date: 2023-11-17, cloud pattern date: 2023-06-26). All these reconstructions are within the sensor's accuracy~\cite{malakar2018operational}. From the reconstruction images and the residuals (Fig.~\ref{fig:ImageClearSkyRealCloud}), we can see that the reconstruction aligns near perfectly with the observed data.

\subsection{Results Under Heavily-Cloudy Situations}
In the second set of experiments, we mainly tested the proposed method under heavily-cloudy situations. In heavily-cloudy situations or even partially-cloudy situations, the valid-observed LST values not covered by clouds should not be considered as clear-sky LST, because the majority of solar energy is reflected back by the heavy clouds compared to the clear-sky situations. The received solar energy is different from a clear-sky condition and depends on the cloud conditions (e.g., percentage of cloud covers, thickness, opacity)~\cite{dickinson1983land}. These valid-observed LST pixels are only observable due to no direct clouds on top of them, but in the next hours they are likely to be blocked when clouds shift a few hundred meters away. Therefore, these heavily-cloudy and even partially-cloudy situations should be treated differently from the clear-sky situations (especially when the percentage of clouds is over 80\%) and should be tested separately. For this purpose, we designed this set of experiments. We selected a few dates that have more than 80\% cloud covers (2023-04-18 for NYC, 2023-05-16 for London, 2023-05-25 for Hong Kong), and used only 80\% of the available valid-observed pixels for training. The remaining 20\% were reserved for testing. 

\begin{figure*}[!t]
   \centering
    \includegraphics[width=0.92\linewidth]{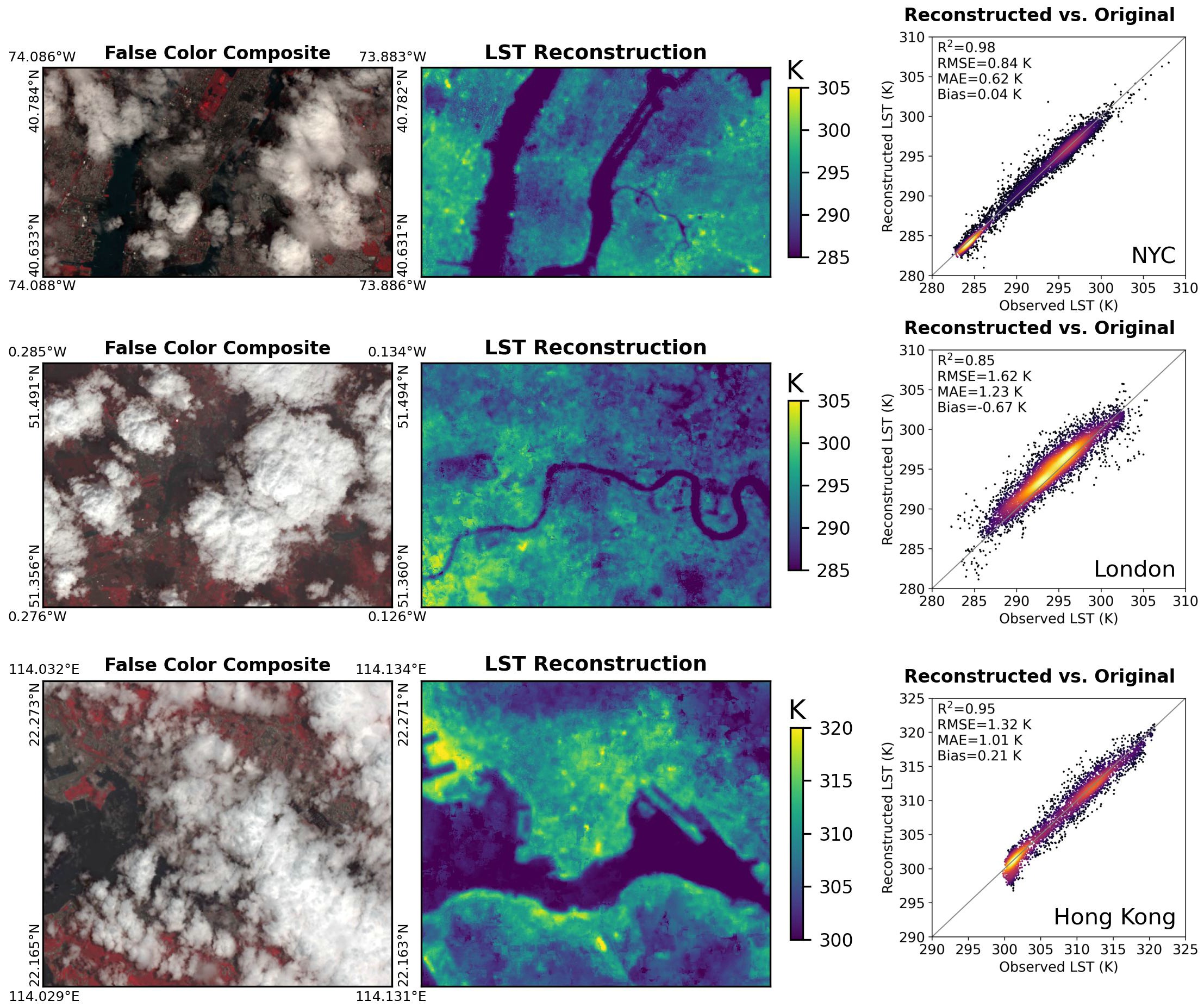}
    \caption{Results under heavily-cloudy situations. From left to right: valid-observed LST, reconstructed LST, scatter plot between reconstructed LST and the 20\% observed LST reserved for testing. From top to bottom: NYC, London, Hong Kong.}
    \label{fig:ResultHeavyCloud}
\end{figure*}

The results are shown in Fig.~\ref{fig:ResultHeavyCloud}. Quantitatively, the proposed method achieved LST reconstruction results of R$^2$=0.85-0.98, RMSE=0.84-1.62 K, MAE=0.62-1.23 K, and Bias=-0.67-0.21 K for the three cities. These reconstructions were obtained with cloud coverage of 83.6\%-85.6\%. Compared to clear-sky situations, the reconstructed LST images under heavily-cloudy situations clearly have some patterns showing the effects of heavy cloud coverage on LST. For example, for the NYC case (Fig.~\ref{fig:ResultHeavyCloud}), the central north areas appear to have relatively lower LST values under heavily-cloudy situations than it is supposed to be under clear-sky situations; and as shown from the cloud patterns in Fig.~\ref{fig:ResultHeavyCloud}, these areas are most affected by clouds. Similarly, the eastern part of London and the central part of Hong Kong (Fig.~\ref{fig:ResultHeavyCloud}) also show relatively lower LST values compared to clear-sky situations. These quantitative results and qualitative visualizations demonstrate the success of the proposed method for LST reconstruction under heavily-cloudy situations. 

\subsection{Indirect Validation via Estimating Near-Surface Air Temperature}
To further validate the proposed method, we tested the performance of the reconstructed LST in estimating near-surface air temperature. To compare the performance of estimating near-surface air temperature through valid-observed LST (under clear-sky situations) and reconstructed LST (under cloud-covered and no Landsat overpass situations), ideally we can apply them together to build one simple linear model, and the results should be consistent regardless of LST situations. But given that there are more cloud-covered and no Landsat overpass data points, the linear model could be dominated by these data points and thus biased. For validation purposes, we should fit two linear models separately, one for clear-sky or valid-observed LST situations, and another for cloud-covered or no Landsat overpass situations. The reconstruction is good if the performance of the latter model has similar capability of the former model. Thus, we built two models using all stations from one city, separately for the two LST situations. We plot the estimated air temperature and in-situ air temperature for each station in Fig.~\ref{fig:resultNYC} for New York City, in Fig.~\ref{fig:resultLDN} for London, and in Fig.~\ref{fig:resultHKG} for selected stations in Hong Kong. Overall, the proposed method achieved similar results on using reconstructed LST to estimate near-surface air temperature (R$^2$=0.93, RMSE=2.11 K, MAE=1.72 K), compared to the estimation directly through valid-observed LST (R$^2$=0.93, RMSE=2.02 K, MAE=1.66 K). 

\subsubsection{New York City}
\begin{figure}[!t]
    \centering
    \includegraphics[width=1.0\linewidth]{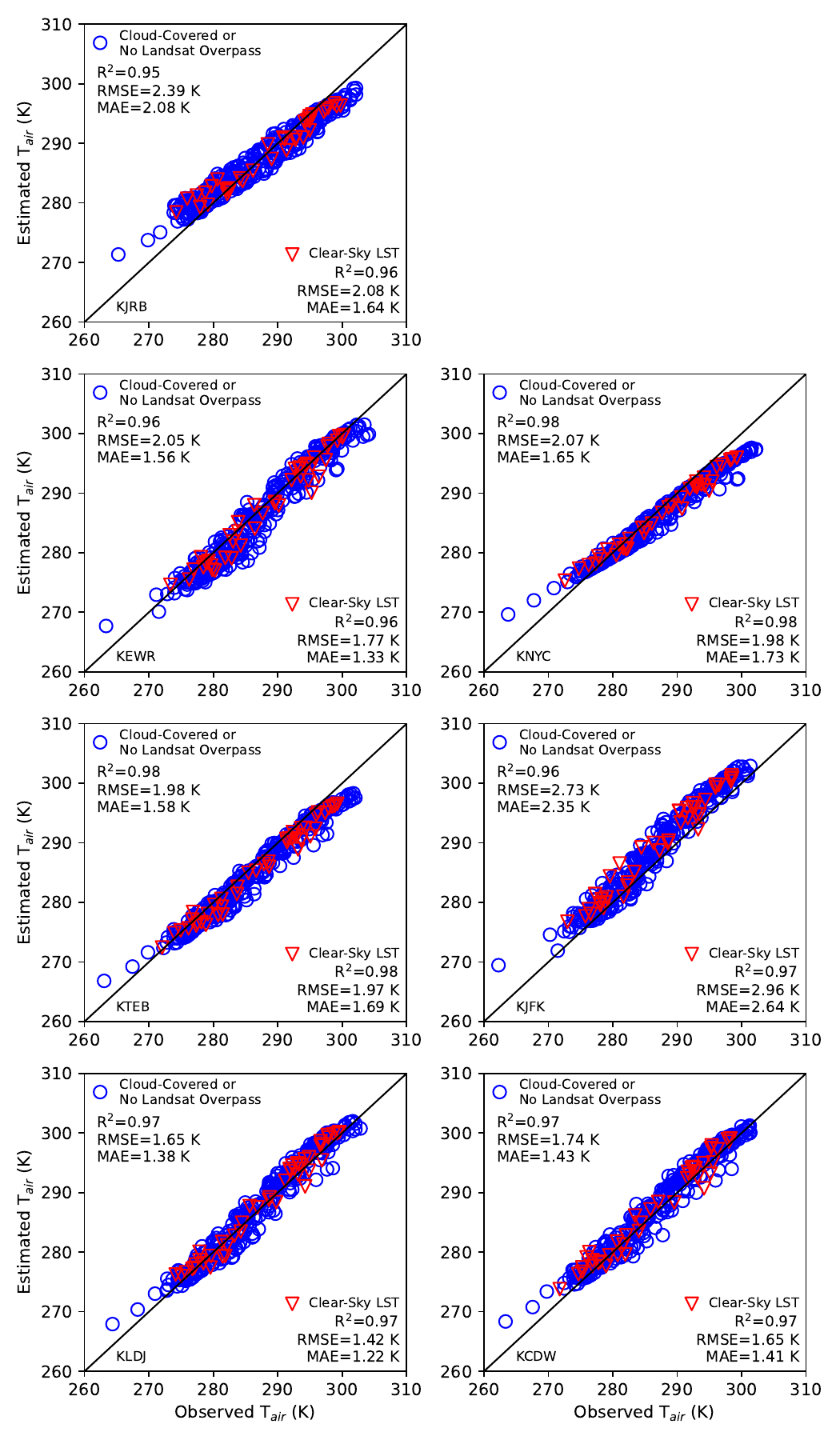}
    \caption{Estimated near-surface air temperature via the proposed method versus in-situ near-surface air temperature from meteorological stations of New York City.}
    \label{fig:resultNYC}
\end{figure}

On the seven meteorological stations in NYC, the estimation of near-surface air temperature achieved comparable results through valid-observed LST (RMSE=1.42-2.96 K) and reconstructed LST (RMSE=1.65-2.73 K). The performance through reconstructed LST is proportionate to the performance through valid-observed LST, meaning that reconstructed LST has consistent characteristics compared with valid-observed LST. From the scatter plots (Fig.~\ref{fig:resultNYC}), we see consistency between the estimated and in-situ near-surface air temperature, except for a few data points at the lower end (below 273.15 K, or 0$^{\circ}$C), which are likely affected by snow. We should note here that this estimation was through very simple linear regression models and by no means the optimized solution to estimate air temperature. We use this method to compare the quality of LST reconstruction only. In summary, the within 2 K estimation for air temperature demonstrates the success of LST reconstruction and its potential in generating accurate air temperature. 

\subsubsection{London}
\begin{figure}[!t]
    \centering
    \includegraphics[width=1.0\linewidth]{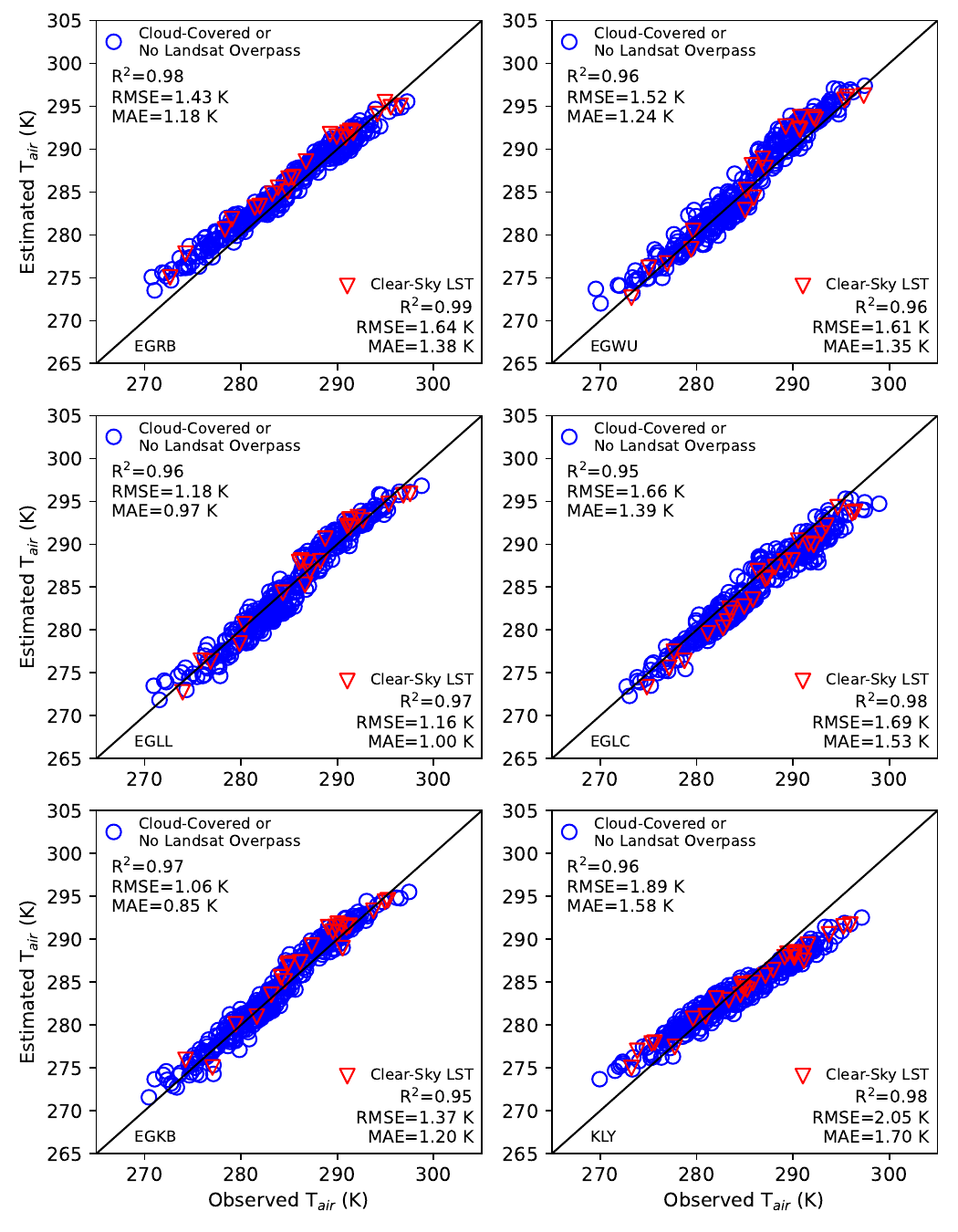}
    \caption{Estimated near-surface air temperature via the proposed method versus in-situ near-surface air temperature from meteorological stations of London.}
    \label{fig:resultLDN}
\end{figure}

For London, the fitting result is excellent, regardless of LST situations (Fig.~\ref{fig:resultLDN}). For valid-observed situations, the fitting result combining all stations is with R$^2$ of 0.93, RMSE of 1.63 K, and MAE of 1.38 K. For each station, the range is R$^2$ of 0.95-0.99, RMSE of 1.16-2.05 K, and MAE of 1.0-1.7 K. For cloud-covered or no Landsat overpass situations, the fitting result combining all stations is R$^2$ of 0.93, RMSE of 1.63, and MAE of of 1.37. For each station, the range is R$^2$ of 0.95-0.98, RMSE of 1.06-1.89, and MAE of 0.85-1.58. The performance through reconstructed LST under cloud-covered or no Landsat overpass situations is indistinguishable from the performance through valid-observed LST. In some cases the performance under cloud-cover situations is even slightly better, which can be attributed to the fact that LST under partially-cloudy and overcast conditions has better agreement with air temperature due to the blocking of direct solar radiation~\cite{vancutsem2010evaluation}. 

\begin{figure}[!t]
    \centering
    \includegraphics[width=1.0\linewidth]{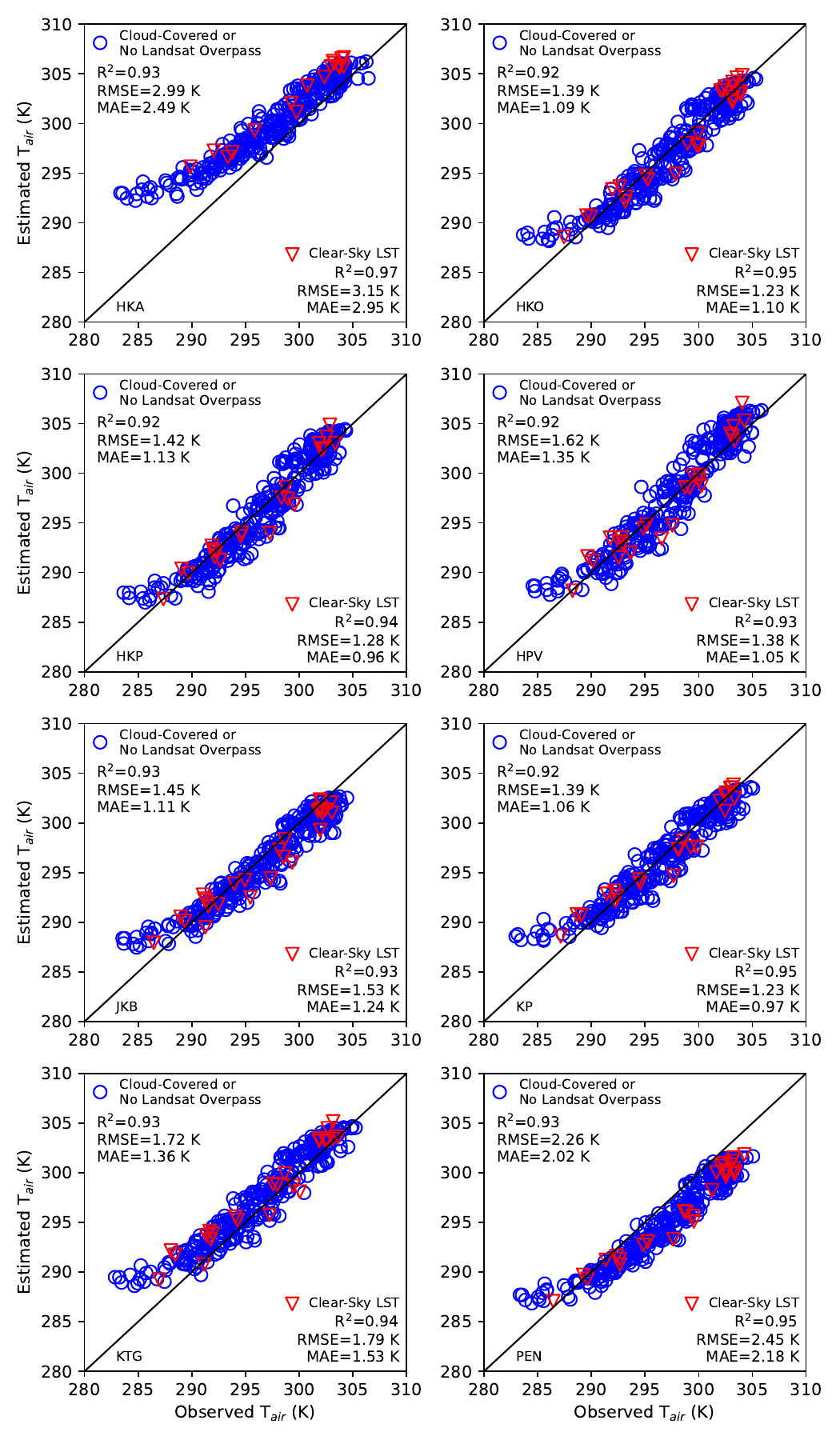}
    \caption{Estimated near-surface air temperature via the proposed method versus in-situ near-surface air temperature from eight selected meteorological stations of Hong Kong.}
    \label{fig:resultHKG}
\end{figure}

\subsubsection{Hong Kong}
We show the scatter plots for the first eight stations (alphabetically) in Fig.~\ref{fig:resultHKG} (the full result from all 23 stations can be found in the supplementary table). Overall, the estimation through reconstructed LST (RMSE=1.73 K) is again indistinguishable compared with the estimation through valid-observed LST (RMSE=1.71 K). The successful estimation of near-surface air temperature in all three cities demonstrates the generalization of the proposed method and indirectly validates the proposed method in LST reconstruction, especially under cloudy situations. 

\begin{figure*}[!t]
\centering
\includegraphics[width=.96\textwidth]{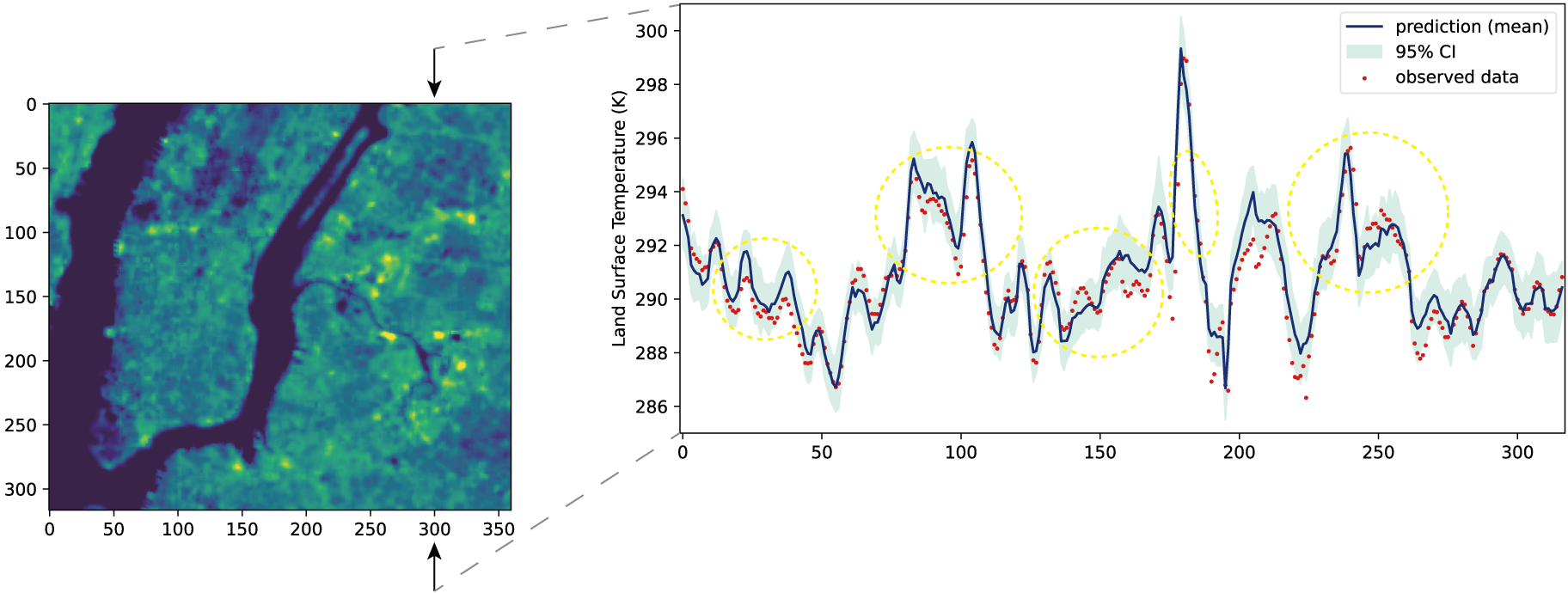}
\caption{Uncertainty Quantification with DELAG along a vertical line at x=300. Landsat observed test data are red dots. Yellow circles highlight reasonable wide or narrow quantified uncertainties.}  
\label{fig:uncertainty}
\end{figure*}

\subsection{Uncertainty Quantification}
The proposed method, DELAG, is the first of its kind that can offer uncertainty quantification. To demonstrate its effectiveness, we visualize the reconstructed data, the 95\% prediction intervals, and the valid-observed Landsat LST values along a vertical line (x-axis=300) in Fig.~\ref{fig:uncertainty}. This particular vertical line is selected because it spans a broad range of values, from 286 K to 300 K, presenting a challenging scenario for uncertainty quantification. Within this area, DELAG achieved a RMSE of 0.568 K, with the 95\% prediction intervals capturing 92.45\% of the test data. Notably, the majority of the Landsat data points (red dots) fall within the 95\%  prediction intervals. 
We highlight five clusters in Fig.~\ref{fig:uncertainty} with yellow circles to demonstrate DELAG's accuracy in estimating the 95\% prediction intervals. Effective uncertainty quantification should feature dynamic confidence intervals that are neither too narrow nor too wide: overly narrow intervals suggest the model underestimates the uncertainty, while overly wide intervals indicate ineffective uncertainty quantification with little to none information. In four of the highlighted circles (1st, 2nd, 3rd, and 5th), certain regions exhibit wide 95\% prediction intervals, accurately encompassing the observed data. In contrast, the 4th yellow circle and the right peak of the 2nd circle show regions with narrow 95\% CI ranges that still successfully capture the observed data. This ability to adaptively adjust the 95\% prediction intervals demonstrates the proposed method's reliability in quantifying uncertainty.

\subsection{Comparison with Existing Methods and Ablation Analysis}
We compare the proposed DELAG with two existing methods: Deep Feature Gaussian Processes (DFGP)~\cite{liu2024deep} and a 4-step ATC-based model (4-step ATC)~\cite{zhu2022reconstruction}, along with three baseline models. The results are shown in Table~\ref{tab:alg_comparison}. The numbers reported here are from the first set of experiments. Across all three datasets, the proposed method (DELAG) consistently outperformed the competitors and the baseline models in almost all metrics.

Compared to the baseline methods, the proposed method DELAG successfully improved the reconstruction, in terms of RMSE, from 1.88~K to 0.73~K for NYC, from 3.33~K to 0.96~K for London, and from 2.50~K to 0.74~K for Hong Kong. The addition of GP greatly improved the results by minimizing bias. The addition of ensemble learning slightly improved the model performance and greatly enhanced its capability to quantify uncertainty, thus providing reliable reconstruction.

We want to further highlight here that only DELAG is a daily LST reconstruction method. DFGP~\cite{liu2024deep} only works when there are valid observations, and the 4-step ATC model~\cite{zhu2022reconstruction} is only effective when there are valid observations within a search window with similar vegetation and land surface. Both models are reduced to ATC only when there are no valid observations. On the contrary, the proposed method, through utilizing ERA5 reanalysis data, yields reasonably good results even under heavily cloudy conditions, as evidence from the previous result on assessing their ability to estimate near-surface air temperature.

\begin{table*}[!t]
  \centering
  \caption{Comparison of algorithms on three datasets using MAE, RMSE, $R^2$, Bias, and 95\% uncertainty coverage (Cov@95). Lower MAE, RMSE, and Bias are better; higher $R^2$ and Cov@95 indicate better performance.}
  \label{tab:alg_comparison}
  \scalebox{0.78}{
  \begin{tabular}{lcccrccccrccccrc}
    \toprule
    \multirow{2}{*}{\textbf{Method}} & 
    \multicolumn{5}{c}{\textbf{NYC}} & 
    \multicolumn{5}{c}{\textbf{London}} & 
    \multicolumn{5}{c}{\textbf{Hong Kong}} \\
    \cmidrule(lr){2-6} \cmidrule(lr){7-11} \cmidrule(lr){12-16}
    & MAE (K) & RMSE (K) & $R^2$ & Bias (K) & Cov@95 & 
      MAE (K) & RMSE (K) & $R^2$ & Bias (K) & Cov@95 & 
      MAE (K) & RMSE (K) & $R^2$ & Bias (K) & Cov@95 \\
    \midrule
    DFGP~\cite{liu2024deep} & 1.25 & 1.69 & 0.83 & -0.24 & 36.32\% & 1.02 & 1.37 & 0.79 & -0.31 & 39.95\% & 1.01 & 1.36 & 0.75 & 0.11 & 38.24\% \\
    4-step ATC~\cite{zhu2022reconstruction} & 0.66 & 0.90 & 0.95 & -0.40 & NA & 2.05 & 2.43 & 0.61 & 1.27 & NA & 0.78 & 1.07 & 0.85 & -0.48 & NA \\
    eATC & 1.55 & 1.88 & 0.93 & 1.31 & NA & 2.88 & 3.33 & 0.51 & 2.56 & NA & 2.06 & 2.50 & 0.73 & 1.92 & NA \\
    eATC$_e$ & 1.39 & 1.65 & 0.96 & 1.31 & 60.53\% & 1.94 & 2.27 & 0.83 & 1.92 & 58.86\% & 1.94 & 2.27 & 0.83 & 1.92 & 58.86\% \\
    eATC + GP & 1.19 & 1.55 & 0.93 & 0.19 & 28.88\% & 1.47 & 1.86 & 0.67 & \textbf{0.01} & 32.18\% & 1.01 & 1.29 & 0.82 & 0.18 & 31.44\% \\
    DELAG & \textbf{0.56} & \textbf{0.73} & \textbf{0.97} & \textbf{-0.01} & \textbf{97.64\%} & \textbf{0.74} & \textbf{0.96} & \textbf{0.90} & -0.27 & \textbf{96.41\%} & \textbf{0.56} & \textbf{0.74} & \textbf{0.93} & \textbf{-0.02} & \textbf{97.95\%} \\
    \bottomrule
  \end{tabular}}
\end{table*}

\subsection{Sensitivity of Features in Gaussian Processes}
We test the sensitivity of using Landsat and Sentinel-2 mean spectral bands as the features used in Gaussian processes, and the results are shown in Table~\ref{tab:LandsatSentinel}. The results from different spectral bands are comparable to each other. Using annual mean spectral bands is the classic option to represent the Earth surface. Recent advances in deep representation learning has shown some promising direction on using features obtained through Earth foundation models for accurate representation~\cite{xiao2024foundation, zhu2024foundations}, which may be integrated here in the future.

\begin{table}[!t]
\centering
\caption{Comparison of Spectral Bands (Landsat vs. Sentinel) for Gaussian Process Construction.}
\label{tab:LandsatSentinel}
\resizebox{\columnwidth}{!}{
\begin{tabular}{llrrr} 
\toprule
\textbf{Study Area} & \textbf{Method}                       & \textbf{MAE (K)} & \textbf{RMSE (K)} & \textbf{R$^2$} \\
\midrule
NYC                 & eATC$_e$                              & 1.39             & 1.65              & 0.96           \\
                    & DELAG (w/ Landsat)                    & 0.55             & 0.72              & 0.97           \\
                    & DELAG (w/ Sentinel)                   & 0.56             & 0.73              & 0.97           \\
\midrule
London              & eATC$_e$                              & 2.75             & 2.90              & 0.84           \\
                    & DELAG (w/ Landsat)                    & 0.67             & 0.89              & 0.91           \\
                    & DELAG (w/ Sentinel)                   & 0.74             & 0.96              & 0.90           \\
\midrule
Hong Kong           & eATC$_e$                              & 1.94             & 2.27              & 0.83           \\
                    & DELAG (w/ Landsat)                    & 0.60             & 0.79              & 0.92           \\
                    & DELAG (w/ Sentinel)                   & 0.56             & 0.74              & 0.93           \\
\bottomrule
\end{tabular}%
} 
\end{table}

\subsection{Impacts of Landsat Data Frequency}
To investigate the impacts of data frequency, we designed a set of experiments with three different observation settings. The three settings are, respectively, dual satellites with Landsat 8 and 9 (4 scenes per 16 days) that is the setting we used in this paper, Landsat 8 with cross-track areas (2 scenes per 16 days) that is a scenario where only one Landsat satellite is available, and Landsat 8 without cross-track areas (1 scene per 16 days). As the numbers of observed samples were not the same across different settings, for fair comparison, we evaluated the performance over their ability to predict near-surface air temperature. For each study area, we combined all data samples (observed or non-observed) together to build one regression model. The results are presented in Table~\ref{tab:DataFrequency}. 

We observe a consistent trend that with more data, the performance in estimating air temperature is better. For NYC and London, with 50\% capacity (2 scenes per 16 days), the performance difference is marginal, with RMSE larger by less than 0.1~K. But with 25\% capacity (1 scene per 16 days), the performance is significantly worse, degrading to 3.03~K RMSE for NYC and 2.71~K RMSE for London. The performance over Hong Kong is significantly degraded with 50\% capacity already, likely due to its more cloudy climate. This set of experiments shows that either cross-track or dual-satellite is important for achieving desirable LST reconstruction using the proposed method.

\begin{table*}[!t]
  \centering
  \caption{Impacts of Landsat data frequency.}
  \label{tab:DataFrequency}
  \begin{tabular}{llccccccccc}
    \toprule
    \multirow{2}{*}{\textbf{Setting}} &
    \multirow{2}{*}{\textbf{Scene Frequency}} & 
    \multicolumn{3}{c}{\textbf{NYC}} & 
    \multicolumn{3}{c}{\textbf{London}} & 
    \multicolumn{3}{c}{\textbf{Hong Kong}} \\
    \cmidrule(lr){3-5} \cmidrule(lr){6-8} \cmidrule(lr){9-11}
    & & MAE (K) & RMSE (K) & $R^2$ & 
      MAE (K) & RMSE (K) & $R^2$ & 
      MAE (K) & RMSE (K) & $R^2$ \\
    \midrule
      Landsat 8 + 9, cross-track    & 4 per 16 days     & 1.66 & 2.06 & 0.94     & 1.11 & 1.39 & 0.94     & 1.32 & 1.72 & 0.89 \\ 
      Landsat 8, cross-track    & 2 per 16 days     & 1.74 & 2.13 & 0.93     & 1.18 & 1.46 & 0.93     & 1.81 & 2.34 & 0.82 \\ 
      Landsat 8, single-track   & 1 per 16 days     & 2.38 & 3.03 & 0.86     & 2.18 & 2.71 & 0.76     & 2.29 & 2.81 & 0.77 \\ 
    \bottomrule
  \end{tabular}
\end{table*}

\section{Discussion and Conclusion}
\label{sec:ConclusionDiscussion}
Having daily temperature data at high spatial resolution is critical for many applications. Yet, existing temperature data, such as gridMET and MODIS, cannot capture the temperature variations within cities. While Landsat is the only satellite series that provides LST data at fine scale (100~m), existing studies have overlooked its potential in generating seamless temperature data. In this study, we leverage the cross-track characteristics and dual-satellite operation of Landsat since 2021 and argue that Landsat is a valuable source for generating seamless temperature data at high spatiotemporal resolution. 

We propose DELAG, a deep ensemble learning method incorporating annual temperature cycles (ATC) and Gaussian processes (GP), for Landsat LST reconstruction. We selected three representative cities---New York City, London and Hong Kong---as study areas to show the generalization of the proposed method. We tested DELAG under three different cloud situations. Under clear-sky situations with real-world cloud patterns, DELAG achieved LST reconstruction with RMSE of 0.73~K, 0.96~K and 0.74~K for the three cities, respectively. Under heavily-cloudy situations, DELAG achieved LST reconstruction with RMSE of 0.84~K, 1.62~K and 1.32~K, respectively. These results are superior to existing methods and, additionally, DELAG can offer uncertainty quantification, providing more reliable reconstructions. We further tested DELAG's capability to estimate near-surface air temperature using simple linear regressions. The proposed method successfully estimated near-surface air temperature with RMSE of 2.11 K, 1.48 K and 1.73 K for the three cities respectively through reconstructed LST, comparable to the estimations through valid-observed LST (RMSE of 2.02~K, 1.63~K and 1.71~K). The successful estimations of near-surface air temperature and the comparable results through reconstructed LST and valid-observed LST indirectly validate DELAG in reconstructing seamless temperature data for real-world applications, such as examining effects of respiratory diseases and wildfire impacts~\cite{liu2024effects, chen2021mortality}. As existing seamless air temperature products are largely derived from satellite observations, the proposed framework and method thus provides a novel and practical pathway for real-world applications requiring seamless temperature data at high spatiotemporal resolution. 

The LST generated from the proposed method is fundamentally an integration of ERA5 signals and observable pixels (either clear-sky or some pixels observable in cloudy conditions). Overcast conditions are not directly observable, so the associated information is estimated (extrapolated) using daily fluctuations from ERA5-a weather and climate reanalysis product-with a linear multiplier. This approach assumes a linear relationship of the variation from mean (after subtracted by ATC) between local pixels and regional temperature values. While this estimation is reasonable and leverages the best available data, it remains an estimation that may not account for all local weather conditions. Nonetheless, the reconstructed LST has demonstrated far more realistic performance than purely clear-sky LST reconstruction, as evidenced by its performance in estimating near-surface air temperature.

Landsat data frequency plays a major role in LST reconstruction. Landsat Next will greatly enhance data frequency, offering better data availability that is critical for daily LST reconstruction~\cite{irons2012next}. Enhancing data availability by integrating multiple LST observations can also enhance daily LST reconstruction globally. Satellites and sensors from the Sentinel satellite series, the GOES satellites, and the International Space Station can be of importance. We plan to integrate multiple satellite sources to achieve daily LST reconstruction in the future. Additionally, the reconstructed data here are still LST instead of air temperature. To fully achieve air temperature mapping, a large number of weather stations and a proper cross-validation strategy to separate training and testing stations are necessary. 

\ifCLASSOPTIONcaptionsoff
  \newpage
\fi

\bibliographystyle{IEEEtran}
\bibliography{refs}

\end{document}